\documentclass{article}

\usepackage[final]{neurips_2020}
\usepackage[utf8]{inputenc}
\usepackage[T1]{fontenc}
\usepackage{hyperref}
\usepackage{url}
\usepackage{booktabs}
\usepackage{amsfonts,amssymb,amsthm,amsmath}
\usepackage{nicefrac}
\usepackage{microtype}

\usepackage{bm}
\usepackage{graphicx}
\usepackage[dvipsnames]{xcolor}
\usepackage[inline]{enumitem}
\usepackage{tikz}
\usepackage{algorithm}
\usepackage[noend]{algpseudocode}
\usepackage{xpatch}
\usepackage{xifthen}

\graphicspath{{figs/}}

\makeatletter
\xpatchcmd{\algorithmic}{\itemsep\z@}{\itemsep=0.5ex plus1pt}{}{}
\makeatother

\setitemize{leftmargin=*}
\setlist{nosep}

\usetikzlibrary{cd,arrows,calc,shapes,positioning}
\tikzstyle{obs} = [circle,fill=white,draw=black,inner sep=1pt,minimum size=20pt,font=\fontsize{10}{10}\selectfont,node distance=1,thick]
\tikzstyle{latent} = [obs,dotted]

\newcommand{\edge}[3][]{ %
  \foreach \x in {#2} { %
    \foreach \y in {#3} { %
      \path (\x) edge [->, >={triangle 45}, #1,thick] (\y) ;%
    } ;
  } ;
}

\definecolor{tab-blue}{rgb}{0.12156862745098039, 0.4666666666666667, 0.7058823529411765}
\definecolor{tab-orange}{rgb}{1.0, 0.4980392156862745, 0.054901960784313725}
\definecolor{tab-green}{rgb}{0.17254901960784313, 0.6274509803921569, 0.17254901960784313}
\definecolor{tab-red}{rgb}{0.8392156862745098, 0.15294117647058825, 0.1568627450980392}
\definecolor{tab-purple}{rgb}{0.5803921568627451, 0.403921568627451, 0.7411764705882353}
\definecolor{tab-cyan}{rgb}{0.09019607843137255, 0.7450980392156863, 0.8117647058823529}

\newcommand{\trueline}{\raisebox{2pt}{\tikz{\draw[tab-red,solid,ultra thick](0, 0) -- (5mm, 0);}}}
\newcommand{\kivline}{\raisebox{2pt}{\tikz{\draw[tab-purple,dashdotted,ultra thick](0, 0) -- (5mm, 0);}}}
\newcommand{\tslsline}{\raisebox{2pt}{\tikz{\draw[tab-cyan,dashed,ultra thick](0, 0) -- (5mm, 0);}}}
\newcommand{\lowerline}{\raisebox{0pt}{%
\tikz{\draw[tab-blue,dotted,ultra thick](0, 0) -- (6mm, 0);%
\draw[tab-blue,line width=1pt](2mm, -1mm) -- +(2mm, 2mm);%
\draw[tab-blue,line width=1pt](2mm, 1mm) -- +(2mm, -2mm);}}}
\newcommand{\upperline}{\raisebox{0pt}{%
\tikz{\draw[tab-orange,dotted,ultra thick](0, 0) -- (6mm, 0);%
\draw[tab-orange,line width=1pt](2mm, -1mm) -- +(2mm, 2mm);%
\draw[tab-orange,line width=1pt](2mm, 1mm) -- +(2mm, -2mm);}}}
\newcommand{\modelline}{\raisebox{2pt}{\tikz{\draw[gray,solid,line width=0.7pt](0, 0) -- (5mm, 0);}}}

\newcommand{\xhdr}[1]{\textbf{#1.}\;}
\newcommand{\bR}{\ensuremath \mathbb{R}}
\newcommand{\bN}{\ensuremath \mathbb{N}}
\newcommand{\cA}{\ensuremath \mathcal{A}}
\newcommand{\cL}{\ensuremath \mathcal{L}}
\newcommand{\cO}{\ensuremath \mathcal{O}}
\newcommand{\cX}{\ensuremath \mathcal{X}}
\newcommand{\cY}{\ensuremath \mathcal{Y}}
\newcommand{\cZ}{\ensuremath \mathcal{Z}}
\newcommand{\cN}{\ensuremath \mathcal{N}}
\newcommand{\cD}{\ensuremath \mathcal{D}}

\DeclareMathOperator*{\argmin}{arg\,min}
\DeclareMathOperator{\E}{\mathbb{E}}
\DeclareMathOperator{\indep}{\perp\!\!\!\perp}
\DeclareMathOperator{\dep}{\not\! \perp\!\!\!\perp}

\newcommand{\B}[1]{\bm{#1}}
\newcommand{\given}{\,|\,}
\newcommand{\lhs}{\mathrm{LHS}}
\newcommand{\rhs}{\mathrm{RHS}}
\newcommand{\abstol}{\epsilon_{\mathrm{abs}}}
\newcommand{\reltol}{\epsilon_{\mathrm{rel}}}
\newcommand{\bin}{\mathrm{bin}}

\title{A Class of Algorithms for \\General Instrumental Variable Models}

\author{%
  Niki Kilbertus\thanks{Majority of work done while at Max Planck Institute for Intelligent Systems and University of Cambridge.} \\
  Helmholtz AI\\
  \And
  Matt~J.~Kusner \\
  University College London \\
  The Alan Turing Institute \\
  \And
  Ricardo Silva \\
  University College London \\
  The Alan Turing Institute \\
}

\begin{document}

\maketitle

\setcounter{footnote}{0}

\begin{abstract}
  Causal treatment effect estimation is a key problem that arises in a variety of real-world settings, from personalized medicine to governmental policy making. There has been a flurry of recent work in machine learning on estimating causal effects when one has access to an instrument. However, to achieve identifiability, they in general require one-size-fits-all assumptions such as an additive error model for the outcome. An alternative is partial identification, which provides bounds on the causal effect. Little exists in terms of bounding methods that can deal with the most general case, where the treatment itself can be continuous. Moreover, bounding methods generally do not allow for a continuum of assumptions on the shape of the causal effect that can smoothly trade off stronger background knowledge for more informative bounds. In this work, we provide a method for causal effect bounding in continuous distributions, leveraging recent advances in gradient-based methods for the optimization of computationally intractable objective functions. We demonstrate on a set of synthetic and real-world data that our bounds capture the causal effect when additive methods fail, providing a useful range of answers compatible with observation as opposed to relying on unwarranted structural assumptions.\footnote{Code available at \url{https://github.com/nikikilbertus/general-iv-models}.}

\end{abstract}

\section{Introduction}
\label{sec:intro}
Machine learning is becoming more and more prevalent in applications that inform actions to be taken in the physical world.
To ensure robust and reliable performance, many settings require an understanding of the causal effects an action will have before it is taken.
Often, the only available source of training data is observational, where the actions of interest were chosen by unknown criteria.
One of the major obstacles to trustworthy causal effect estimation with observational data is the reliance on the strong, untestable assumption of \emph{no unobserved confounding}.
To avoid this, only in very specific settings (e.g., front-door adjustment, linear/additive instrumental variable regression) it is possible to allow for unobserved confounding and still identify the causal effect \citep{pearl2009causality}.
Outside of these settings, one can only hope to meaningfully bound the causal effect \citep{manski:07}.

In many applications, we have one or few treatment variables $X$ and one outcome variable $Y$.
Nearly all existing approaches to obtain meaningful bounds on the causal effect of $X$ on $Y$ impose constraints on how observed variables are related, in order to mitigate the influence of unobserved confounders.
One of the most useful structural constraints is the existence of an observable \emph{instrumental variable} (IV):
a variable $Z$, not caused by $X$, whose relationship with $Y$ is entirely mediated by $X$, see \citet{pearl2009causality} for a graphical characterization. 
The existence of an IV can be used to derive upper (lower) bounds on causal effects of interest by maximizing (minimizing) those effects among all IV models compatible with the observable distribution.
\emph{In this work, we develop algorithms to compute these bounds on causal effects over ``all'' IV models compatible with the data in a general continuous setting.}
Crucially, the space of ``all'' models \emph{cannot} be arbitrary, but it can be made very flexible.
Instead of forcing a user to adopt a model space with hard constraints, we will allow for choice from a continuum of model spaces.
Our approach rewards background knowledge with tighter bounds and it is not tied to an a priori inflexible choice, such as additivity or monotonicity.
It avoids the adoption of unwarranted structural assumptions under the premise that they are needed due to the lack of ways of expressing more refined domain knowledge.
The burden of the trade-off is put explicitly on the practitioner, as opposed to embracing possibly crude approximations due to the limitations of identification strategies.

Eliciting constraints that characterize ``the models compatible with data'' under a causal directed acyclic graph (DAG) for discrete variables is an active field of study, with contributions from the machine learning, algebraic statistics, economics, and quantum mechanics literature.
This has provided complete characterizations of equality \citep{evans:19,tian:02} and inequality \citep{wolfe2019inflation,wolfe2019completeness} constraints.
Enumerating all inequality constraints is in general super-exponential in the number of observed variables, even for discrete causal models. 
However, this line of work typically solves a harder problem than is strictly required for bounding causal effects: they provide symbolic constraints obtained by eliminating all hidden variables.
While the pioneering work of \cite{balke1994counterfactual} in the discrete setting also provides symbolic constraints via a super-exponential algorithm, it introduces constraints that match the observed marginals of a latent variable model against the observable distribution.
Thereby it provides a connection to non-symbolic, stochastic approaches for evaluating integrals, which we develop in this work.

Our key observation is that we can leverage recent advances in efficient gradient and Monte Carlo-based optimization of computationally intractable objective functions to bound the causal effect directly.
This can be done even in the setting where $X$ is continuous, where none of the literature described above applies.
We do so by (a) parameterizing the space of causal responses to treatment $X$ such that we can incorporate further assumptions that lead to informative bounds; (b) using a Monte Carlo approximation to the integral over the distribution of possible responses to $X$, where the distribution itself must be parameterized carefully to incorporate the structural constraints of an IV DAG model.
This allows us to optimize over the domain-dependent set of all plausible models that are consistent with observed data to find lower/upper bounds on the target causal effect.

In Section \ref{sec:background}, we describe the general problem of using instrumental variables when treatment $X$ is continuous.
Section \ref{sec:setting} develops our representation of the causal model.
In Section \ref{sec:optimization} we introduce a class of algorithms for solving the bounding problem and our suggested implementation.
Section \ref{sec:results} provides several demonstrations of the advantages of our method.

\section{Current Approaches and Their Limitations}
\label{sec:background}
\citet{balke1994counterfactual} focused on partial identification (bounding) of causal effects on binary discrete models.
\citet{angrist1996identification} studied identification of effects for a particular latent subclass of individuals also in the binary case.
Meanwhile, the econometrics literature has focused on problems where the treatment $X$ is continuous \citep{newey2003instrumental,blundell2007semi,angrist2008mostly,wooldridge2010econometric,darolles2011nonparametric,horowitz2011applied,chen2018optimal,lewis:18}.
This problem has recently received attention in machine learning, using techniques from deep learning \citep{hartford2017deep,bennet:19} and kernel machines \citep{singh2019kernel,muandet2019dual}.
This literature assumes that the structural equation for $Y$ has a special form, such as having an additive error term $e_Y$, as in $Y = f(X) + e_Y$.
The error term $e_Y$ is not caused by $X$, but need not be independent of it, introducing unobserved confounding.
This assumption is also used in related contexts, such as in sensitivity analysis for counterfactual estimands, see \citet{kilbertus2019sensitivity} for a specific application in fairness.

Using the notation of \cite{pearl2009causality}, the expected response under an intervention on $X$ at level $x$ is denoted by $\E[Y\given do(x)]$, which in the model above boils down to $f(x)$.
An \emph{average treatment effect} (ATE) can be defined as a contrast of this expected response under two treatment levels, e.g., $f(x) - f(x')$.
In the zero-mean additive error case, $\E[Y \given z] = \int f(x) p(x \given z)\, dx$.
Under some regularity conditions, no function other than $f(\cdot)$ satisfies that integral equation.
Since $\E[Y \given z]$ and $p(x \given z)$ can be both learned from data, this allows us to learn the ATE from observational data.
This is how the vast majority of recent work identifies the causal treatment effect in the IV model \citep{hartford2017deep,bennet:19,singh2019kernel,muandet2019dual}.

The price paid for identification is that it seriously limits the applicability of these models.
Diagnostic tests for the additivity assumption are not possible, as residuals $Y - f(X)$ can be arbitrarily associated with $X$ by assumption.
On the other hand, without \emph{any} restrictions on the structural equations, it is not only impossible to identify the causal effect of the IV model with a continuous treatment, but even bounds on the ATE are vacuous \citep{pearl1995testability,bonet:01,gunsilius2018testability,gunsilius2019bounds}.
However, with relatively weak assumptions on the space of allowed structural equations, it is possible to achieve meaningful bounds on the causal effect \citep{gunsilius2019bounds}.
It suffices that the equations for $X$ and $Y$ have a finite number of discontinuities.
Gunsilius provides a theoretical framework for representation and estimation of bounds.
Algorithmically, he proposes a truncated wavelet representation for the causal response and builds convex combinations of a sample of response functions to optimize IV bounds.
Although it is an important proof of concept for the possibility of bounds for the general IV case with a strong theoretical motivation, we found that the method has frequent stability issues that are not easy to diagnose.
We return to this in Appendix~\ref{app:gunsilius}.

Building on top of this work and some classical ideas first outlined by \cite{balke1994counterfactual}, we propose an alternative formulation for finding bounds when both $X$ and $Y$ are continuous.
Our technique flexibly parameterizes the causal response functions, while naturally encoding the structural IV constraints for compatibility with the observed data.
We then leverage an augmented Lagrangian method that is tailored to non-convex optimization with inequality constraints.
We demonstrate that our method matches estimation results of prior work in the additive setting, and gives meaningful bounds on the causal effect in general, non-additive models.
Thereby, we follow a line of recent successes in various domains achieved by replacing previous intractable symbolic-combinatorial algorithms \citep{balke1994counterfactual,wolfe2019inflation,drton:09} with a continuous program.
One of our key contributions is to formulate bounds on true causal effects as well as their compatibility requirements as a smooth, constrained objective, for which we can leverage efficient gradient-based optimization techniques with Monte Carlo approximations.

\section{Problem Setting}
\label{sec:setting}

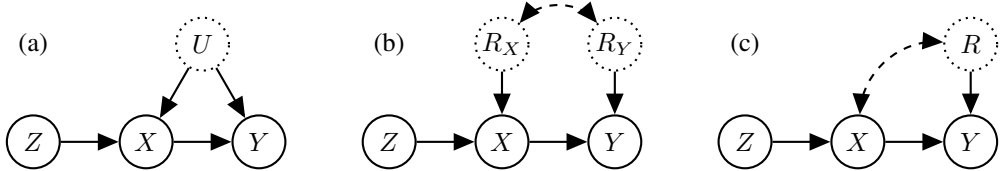
\begin{figure}
  \centering
  \begin{tikzpicture}
  \node (L) at (0, 1.3) {(a)};
  \node[obs] (Z) at (0,0) {$Z$};
  \node[obs] (X) at (1.5, 0) {$X$};
  \node[obs] (Y) at (3, 0) {$Y$};
  \node[latent] (U) at (2.25, 1.3) {$U$};
  \edge{Z}{X};
  \edge{X}{Y};
  \edge{U}{X,Y};
  \end{tikzpicture}%
  \hspace{1cm}%
  \begin{tikzpicture}
  \node (L) at (0, 1.3) {(b)};
  \node[obs] (Z) at (0,0) {$Z$};
  \node[obs] (X) at (1.5, 0) {$X$};
  \node[obs] (Y) at (3, 0) {$Y$};
  \node[latent] (rY) at (3, 1.3) {$R_Y$};
  \node[latent] (rX) at (1.5, 1.3) {$R_X$};
  \edge{Z}{X};
  \edge{X}{Y};
  \edge{rX}{X};
  \edge{rY}{Y};
  \edge[<->, bend left=50, dashed] {rX}{rY};
  \end{tikzpicture}%
  \hspace{1cm}%
  \begin{tikzpicture}
  \node (L) at (0, 1.3) {(c)};
  \node[obs] (Z) at (0,0) {$Z$};
  \node[obs] (X) at (1.5, 0) {$X$};
  \node[obs] (Y) at (3, 0) {$Y$};
  \node[latent] (rY) at (3, 1.3) {$R$};
  \edge{Z}{X};
  \edge{X}{Y};
  \edge{rY}{Y};
  \edge[<->, bend left=45, dashed] {X}{rY};
  \end{tikzpicture}%
 \caption{(a) An example of DAG compatible with $Z$ being an instrument for $X \rightarrow Y$, with hidden confounder $U$.
 (b) An equivalent representation using \emph{response function} indices for deterministic functions $X = g_{R_X}(Z)$ and $Y = f_{R_Y}(X)$, with two random indexing variables $R_X$ and $R_Y$.
 (c) For the purposes of modeling $\E[Y \given do(x)]$, it is enough to express the model in terms of $R := R_Y$ only.}
  \label{fig:setup}
\end{figure}

Following Pearl's Structural Causal Model (SCM) framework \citep{pearl2009causality}, we assume the existence of structural equations and a (possibly infinite dimensional) unobserved exogenous process $U$,
\begin{equation}\label{eq:structural_equations}
  X = g(Z, U) \quad \text{and} \quad Y = f(X, U).
\end{equation}
We illustrate this situation in Figure~\ref{fig:setup}(a).
It assumes the usual requirements for the instrument $Z$ to be satisfied, namely (a) $Z \indep U$, (b) $Z \dep X$, and (c) $Z \indep Y \given \{X, U\}$.

\subsection{Goal}
\label{subsec:setup}

The goal is to compute lower/upper bounds on $\E[Y \given do(x^\star)]$ for any desired intervention level $x^\star$.
Bounds on (conditional) ATEs can be derived, see also Appendix \ref{app:p_eta}.
Intuitively, we put bounds on how $f(X, U)$ depends on $X$ by optimizing over ``allowed'' distributions of $U$.
Which distributions are ``allowed'' is determined by observations, i.e., we only consider settings where marginalizing $U$ results in $p(x, y \given z)$ for all $(x, y, z)$ in the support of the observational distribution.
In fact, as pointed out by \cite{palmer:11}, it is enough to consider matching the marginals of the latent variable model to the two conditional densities $p(x\given z)$ and $p(y \given z)$\footnote{In Appendix \ref{app:p_yxz}, we discuss the case where we match $p(y \given x, z)$, which can further tighten bounds with some computational advantages and disadvantages compared to $p(y \given z)$.}.
Informally, \emph{among all possible structural equations $\{g, f\}$ and distributions over $U$ that reproduce the estimated densities $\{\hat{p}(x \given z), \hat{p}(y \given z)\}$, we find estimates of the minimum and maximum expected outcomes under intervention}.

\xhdr{Response functions}
The main idea of \cite{balke1994counterfactual} is to express structural equations in terms of \emph{response functions}: labeling (and possibly clustering) states of $U$ according to the implied functional relationship between the observed variable and its direct causes.
These $U$ states are mapped to a particular level of an index variable $R$.
For instance, if $Y = f(X, U) = \lambda_1X + \lambda_2XU_1 + U_2$, a two-dimensional $U$ space in a linear, non-additive outcome function, we have that $f(x, u) = \lambda_1x + \lambda_2x$ for $u_1 = 1$, $u_2 = 0$.
We can define an implicit arbitrary value $r$ such that $f_r(x) = \lambda_rx$, $\lambda_r = \lambda_1 + \lambda_2$, the value ``$r$'' being an alias for $(1, 0)$ in the space of the confounders.
The advantage of this representation is that we can think of a distribution over $R$ as a distribution over functions of $X$ alone.
Otherwise we would need to deal with interactions between $U$ and $X$ on top of a distribution over $U$, itself of unclear dimensionality.
In contrast, the dimensionality of $R$ is the one implied by the particular function space adopted.
\cite{gunsilius2019bounds} provides a more thorough discussion of the role of response functions corresponding to a possibly infinite-dimensional $U$.
Figure~\ref{fig:setup}(b) shows a graphical representation of a system parameterized by response function indices $R_X$ and $R_Y$, with a bi-directed edge indicating possible association between the two.
In what follows, as there will be no explicit need for $R_X$, the causal DAG corresponding to our counterfactual model is shown in Figure~\ref{fig:setup}(c)\footnote{It is also possible to represent only $R_X$ and drop $R_Y$.
\citet{zhang:20} do this in a way that provides a new view of the discrete treatment case.}.
This itself departs from \cite{balke1994counterfactual} and \cite{gunsilius2019bounds}, having the advantage of simplifying the optimization and not assuming counterfactuals for $X$ (which will not exist if $Z$ is not a cause of $X$ but just confounded with it).
Furthermore, focusing on $\{p(x \given z), p(y \given z)\}$ instead of $p(x, y \given z)$ does not require simultaneous measurements of $X$ and $Y$ \citep{palmer:11}, see Appendix \ref{app:p_yxz} for the latter.
Within this framework, we can rewrite the optimization over allowed distributions of $U$ into an optimization over allowed distributions of response functions for $Y$.

Without restrictions on the function space, non-trivial inference is impossible \citep{pearl1995testability,bonet:01,gunsilius2018testability}.
In our proposed class of solutions, we will adopt a parametric response function space: each response type $r$ corresponds to some parameter value $\theta_r \in \Theta \subset \bR^K$ for some finite $K$.
We write $f_{r}(x) := f_{\theta_r}(x)$.
Going forward, we will simply use $\theta$ to denote a specific response type and drop the index $r$.
While our method works for any differentiable $f_{\theta}$, we will focus on linear combinations of a set of basis functions $\{\psi_k: \bR \to \bR\}_{k \in [K]}$\footnote{We use the notation $[K] := \{1, \ldots, K\}$ for $K \in \bN_{>0}$.} with coefficients $\theta \in \Theta$:
\begin{equation}\label{eq:response_func}
  f_{\theta}(x) := \sum_{k=1}^{K}\, \theta_k\, \psi_k(x).
\end{equation}

We propose to optimize over distributions $p_{\mathcal{M}}(\theta)$ of the response function parameters $\theta$ in the unknown causal model $\mathcal{M}$, subject to the observed marginal of the model, $\int p_{\mathcal{M}}(x, y \given z, \theta)p_{\mathcal{M}}(\theta)\,d\theta$, matching the corresponding (estimated) marginals $p(y\given z)$ and $p(x \given z)$.
Notice that $\theta \indep Z$ is implied by $Z \indep U$ in the original formulation in terms of exogenous variables $U$.
We assume a parametric form for $p_{\mathcal{M}}(\theta)$ via parameters $\eta \in \bR^d$, denoted by $p_{\eta}(\theta)$.
We propose to use function families for $p_{\eta}(\theta)$ that allow for practically low-variance Monte-Carlo gradient estimation via the reparameterization trick \citep{kingma2013auto} to learn $\eta$ --- more in Section~\ref{subsec:preserveXZ}. 

\xhdr{Objective}
An upper bound for the expected outcome under intervention can be directly written as
\begin{equation}\label{eq:objective}
  \max_{\eta} \E[Y \given do(x^{\star})] = \max_{\eta} \int f_{\theta}(x^{\star}) \, p_{\eta}(\theta)\, d\theta.
\end{equation}
A lower bound can be found analogously by the minimization problem.
When optimizing eq.~\eqref{eq:objective} constrained by $p(y \given z)$ and $p(x \given z)$ in the sequel, it will be necessary to define $p_\eta(x, \theta \given z)$.\footnote{We abuse notation slightly by expanding the definition of $\eta$ to simultaneously signify all parameters specifying this joint distribution, as well as individual parameters specific to certain factors of the joint.}
In particular, $\int p_\eta(x, \theta \given z)\,dx = p_\eta(\theta \given z) = p_\eta(\theta)$.
The last equality will be enforced in the encoding of $p_\eta(x, \theta \given z)$, as we need $Z \indep \theta$ even if $Z \dep \theta \given X$.
This encoding is introduced in Section~\ref{subsec:preserveXZ}, which will also allow us to easily match the marginal $p(x \given z)$.
In Section~\ref{subsec:preserveYZ}, we construct constraints for the optimization so that the marginal of $Y$ given $Z$ in $\mathcal{M}$ matches the model-free $p(y \given z)$.

\subsection{Matching \texorpdfstring{$p(x \given z)$}{p(x | z)} and Enforcing \texorpdfstring{$Z \indep U$}{Z independent of U}}\label{subsec:preserveXZ}

Instead of formulating the criterion of preserving the observed marginal $p(x \given z)$ as a constraint in the optimization problem, we bake it directly into our model.\footnote{A full discussion on the construction and implications of such assumptions is given in Appendix \ref{app:p_eta}.}
To accomplish that, we factor $p_\eta(x, \theta \given z)$ as $p(x \given z)\, p_\eta(\theta \given x, z)$.
The first factor is identified from the observed data and we can thus force our model to match it.
The second factor must be constructed so as to enforce marginal independence between $\theta$ and $Z$ (as required by $Z \indep U$).
We achieve that by parameterizing it by a copula density $c_\eta(\cdot)$ that takes univariate CDFs $F(\cdot)$, which uniquely define the distributions, as inputs,
\begin{equation}\label{eq:copula}
  p_{\eta}(\theta \given x, z) := c_{\eta}(F(x \given z), F_{\eta}(\theta_1), \ldots, F_{\eta}(\theta_K))\prod_{k=1}^K p_{\eta}(\theta_k).
\end{equation}
Here we assume that each component $\theta_k$ of $\theta$ has a Gaussian marginal density with mean $\mu_k$ and variance $\sigma_k^2$, i.e., $p_{\eta}(\theta_k) = \cN(\theta_k; \mu_k, \sigma_k^2)$.
Moreover, assuming $c_{\eta}$ is a multivariate Gaussian copula density requires a correlation matrix $S \in \bR^{(K+1) \times (K+1)}$ for which we only keep a Cholesky factor $L$ without further constraints, rescaling $L^\mathsf{T}L$ to have a diagonal of 1s.
Our full set of parameters is
\begin{equation*}
  \eta := \{\mu_1, \ln(\sigma_1^2), \dots, \mu_K, \ln(\sigma_K^2), L\} \in \bR^{K(K+1) / 2 + 2 K}.
\end{equation*}

\subsection{Matching \texorpdfstring{$p(y \given z)$}{p(y | z)}}
\label{subsec:preserveYZ}

In the continuous output case, our parameterization implies the following set of integral equations
\begin{equation}\label{eq:constraints_exact}
  \Pr(Y \leq y \given Z = z) = \int \B{1}(f_{\theta}(x) \leq y) \, p_\eta(x, \theta \given z)\,dx~d\theta,
\end{equation}
for all $y \in \cY, z \in \cZ$, the respective sample spaces of $Y$ and $Z$, where $\B{1}(\cdot)$ is the indicator function.
These constraints immediately introduce two difficulties.
First, we have an infinite number of constraints to satisfy.
Second, the right-hand side involves integrating non-continuous indicator functions, which poses a problem for smooth gradient-based optimization with respect to $\eta$.\footnote{We discuss discrete outcomes or discrete features, which could also lead to discontinuous $f_{\theta}$ in Appendix~\ref{app:discrete_outcomes}.}

To circumvent these issues, we first choose a finite grid $\{z^{(m)}\}_{m=1}^M \subset \cZ$ of size $M \in \bN$, instead of conditioning on all values in $\cZ$. 
We compute $z^{(m)}$ from a uniform grid on the CDF $F_Z$ of $Z$, i.e., $z^{(m)} := F_Z^{-1}(\nicefrac{m}{M + 1})$ for $m \in [M]$.
Second, to avoid the integration of non-continuous indicator functions, we can express the constraints of eq.~\eqref{eq:constraints_exact} in terms of expectations over a dictionary of $L$ basis functions $\{\phi_l\}_{l=1}^L$.
This leads to the following constraints for $p(y \given z)$:
\begin{equation}\label{eq:constraints_approx2}
  \E[\phi_l(Y) \given z^{(m)}] = \int \phi_l(f_{\theta}(x)) \, p_\eta(x, \theta \given z^{(m)})\, dx~d\theta \quad \text{for all } l \in [L], m \in [M].
\end{equation}
This idea borrows from mean embeddings, where one can reconstruct $p(y \given z)$ from an infinite dictionary sampled at infinitely many points in $\cZ$ \citep{singh2019kernel}.
In this work, we choose an even simpler approach and only constrain moments like mean and variance $\phi_1(Y) := \E[Y]$, $\phi_2(Y) := \mathbb{V}[Y]$, \ldots{}. 
Crucially, we note that \emph{our approximations can only relax the constraints}, i.e., the optima may result in looser bounds compared to the full constraint set, \emph{but not invalid bounds}, barring bad local optima as well as Monte Carlo and estimation errors.

\section{Optimization Strategy}
\label{sec:optimization}
Here we state our final non-convex, yet smooth, constrained optimization problem:
\begin{align*}
  \text{objective:} \quad && o_{x^{\star}}(\eta) &:= \int f_{\theta}(x^{\star}) \, p_{\eta}(\theta)\, d\theta \\[-1.5mm]
  \text{constraint LHS:} \quad &&\lhs_{m, l} &:= \E[\phi_l(Y) \given z^{(m)}] \\[-1.5mm]
  \text{constraint RHS:} \quad &&\rhs_{m, l}(\eta) &:= \int \phi_l(f_{\theta}(x))\, p_{\eta} (x, \theta \given z^{(m)} ) \, dx\, d\theta\\[-1.5mm]
  \text{\textbf{opt.\ problem:}} \quad && \min_{\eta} / \max_{\eta} o_{x^{\star}}(\eta) \quad &\text{s.t.} \quad \lhs_{m, l} = \rhs_{m, l}(\eta) \; \text{for all } m \in [M], l\in [L]
\end{align*}
Here, $\min$ and $\max$ give the lower and upper bound respectively.
In this section we describe how to tackle the optimization with an augmented Lagrangian strategy \citep{nocedal2006numerical} and how to estimate all quantities from observed data.
Algorithm~\ref{alg:optimization} in Appendix~\ref{app:algo} describes the full procedure.

\subsection{Augmented Lagrangian Strategy}

We can think of the left-hand side $\lhs$ as target values, estimated once up front from observed data.
The right-hand side $\rhs$ is estimated repeatedly using (samples from) our model $p_{\eta}(x, \theta \given z^{(m)})$ during optimization.
For notational simplicity, we will often ``flatten'' the indices $m$ and $l$ into a single index $l \in [M\cdot L]$.
Since $\lhs$ is subject to misspecification and estimation error, we introduce positive tolerance variables $b \in \bR^{M \cdot L}_{>0}$, relaxing equality constraints into inequality constraints
\begin{equation*}
  c_l(\eta) := b_l - |\lhs_l - \rhs_l(\eta)| \ge 0, \quad \text{with }
  b_l := \max \{\abstol, \reltol \cdot |\lhs_l|\},
\end{equation*}
for fixed absolute and relative tolerances $\abstol,\reltol > 0$.
The constraint $c_l(\eta)$ is satisfied if $\rhs_l(\eta)$ is \emph{either} within a fraction $\reltol$ of $\lhs_l$ \emph{or} within $\abstol$ of $\lhs_l$ in absolute difference.
The absolute tolerance is useful when $\lhs$ is close to zero.
The exact constraints are recovered as $\abstol, \reltol \to 0$.
Again, the introduced tolerance can only make the obtained bounds looser, not invalid.

We consider an inequality-constrained version of the augmented Lagrangian approach with Lagrange multipliers $\lambda \in \bR^{M \cdot L}$ (detailed in Section 17.4 of \citet{nocedal2006numerical}). Specifically, the Lagrangian we aim to minimize with respect to $\eta$ is:
\begin{equation}\label{eq:subproblem}
  \cL(\eta, \lambda, \tau) := \pm o_{x^{\star}}(\eta) + \sum_{l=1}^{M \cdot L}
  \begin{cases}
    - \lambda_l c_l(\eta) + \frac{\tau c_l(\eta)^2}{2}
      & \text{if } \tau c_l(\eta) \le \lambda_l,  \\
    - \frac{\lambda_l^2}{2 \tau}
      & \text{otherwise},
  \end{cases}
\end{equation}
where $+$/$-$ is used for the lower/upper bound and $\tau$ is a temperature parameter, which is increased throughout the optimization procedure.
Given an approximate minimum $\eta$ of this subproblem, we then update $\lambda$ and $\tau$ according to $\lambda_l \gets \max\{0, \lambda_l - \tau c_l(\eta)\}$ and $\tau \gets \alpha \cdot \tau$ for all $l \in [M \cdot L]$ and a fixed $\alpha > 1$.
The overall strategy is to iterate between minimizing eq.~\eqref{eq:subproblem} and updating $\lambda_l$ and $\tau$.

\subsection{Empirical Estimation and Implementation Choices}

For a dataset $\cD = \{(z_i, x_i, y_i)\}_{i=1}^N \subset \bR^3$, we describe our method in Algorithm~\ref{alg:optimization} in Appendix~\ref{app:algo}.

\xhdr{Pre-processing}
As a first step, we whiten the data (subtract mean, divide by variance).
Then, we interpolate the CDF $\hat{F}_Z$ of $\{z_i\}_{i=1}^N$ to compute the grid points $z^{(m)}$.
Next, we assign each observation to a grid point via $\bin(i) := \max\{\argmin_{m \in [M]} |z_i - z^{(m)}|\}$ for $i \in [N]$, i.e., each datapoint is assigned to the gridpoint that is closest to its $z$-value (higher bin for ties).
Given $M, L$ and $\phi_l$, we can estimate $\lhs_{m,l}$ from data via $\lhs_{m, l} := \E[\phi_l(Y) \given z^{(m)}] \approx \frac{1}{|\bin^{-1}(m)|} \sum_{i \in \bin^{-1}(m)} \phi_l(y_i)$, which remain unchanged throughout the optimization.
This allows us to fix the tolerances $b = \max\{\abstol, \reltol \, \lhs\}$.
Finally, we obtain a single batch of examples from $X \given z^{(m)}$ of size $B \in \bN$, which we will also reuse throughout the optimization via inverse CDF sampling $\hat{x}_j^{(m)} = \hat{F}_{X\given z^{(m)}}^{-1}(\nicefrac{j-1}{B-1})$ for $j \in [B], m \in [M]$.
Here, $\hat{F}_{X\given z^{(m)}}$ is the CDF of $\{x_i\}_{i \in \bin^{-1}(m)}$.

\xhdr{Monte Carlo estimation\footnote{To be clear, \emph{for the particular choice of $\phi$ and $p_\eta(\cdot \given \cdot)$ in our experiments, Monte Carlo is not necessary}, see Appendix \ref{app:p_yxz}.  All experiments are done using Monte Carlo to test its suitability for general use.}}
To minimize the Lagrangian, we use stochastic gradient descent (SGD).
Therefore, we need to compute (estimates for) $\nabla_{\eta}\, o_{x^{\star}}(\eta), \nabla_{\eta}\, c_l(\eta)$, where the latter boils down to $\nabla_{\eta}\, \rhs_{m,l}(\eta)$.
In practice, we compute Monte Carlo estimates of $o_{x^{\star}}(\eta)$ and $\rhs_{m,l}(\eta)$ and use automatic differentiation, e.g., using JAX \citep{jax2018github}, to get the gradients.
If we had a batch of independent samples $\theta^{(j)} \sim p_{\eta}(\theta)$ of size $B$, we could estimate the objective eq.~\eqref{eq:objective} for a given $\eta$ via $\E[Y \given do(x^{\star})] \approx \frac{1}{B} \sum_{j = 1}^B f_{\theta^{(j)}}(x^{\star})$.
Similarly, with i.i.d.\ samples $\theta^{(j)} \sim p_{\eta}(\theta \given z^{(m)})$ we can estimate $\rhs_{m, l}$ in eq.~\eqref{eq:constraints_approx2} as $\rhs_{m,l}(\eta) \approx \frac{1}{B} \sum_{j = 1}^B \phi_l\bigl(f_{\theta^{(j)}}(\hat{x}^{(m)}_j)\bigr)$.
Hence, the last missing piece is to sample from eq.~\eqref{eq:copula} in a fashion that maintains differentiability w.r.t.\ $\eta$.
We follow the standard procedure to sample from a Gaussian copula for the parameters $\theta^{(j)}$, with the additional restriction to preserve the pre-computed sample $\hat{x}$.
Algorithm~\ref{alg:sampling} in Appendix~\ref{app:algo} describes the sampling process from $p_{\eta}(\theta, X \given z^{(m)})$ as defined in Section~\ref{subsec:preserveXZ} in detail.
The output is a $(K+1) \times B$-matrix, where the first row contains $B$ independent $X$-samples and the remaining $K$ rows are the components of $\theta \in \bR^{K}$.
We pool samples from all $z^{(m)}$ to obtain samples from $p_{\eta}(\theta)$.
By change of variables, the parameters $\eta = (\mu, \sigma^2, L)$ enter in a differentiable fashion (c.f.\ reparameterization trick \citep{kingma2013auto}).
We initialize $\eta$ randomly, described in detail in Appendix~\ref{app:initialization}.

\xhdr{Response functions}
For the family of response functions we first consider polynomials, i.e., $\psi_k(x) = x^{k - 1}$ for $k \in [K]$.
We will specifically focus on linear ($K = 2$), quadratic ($K = 3$), and cubic ($K = 4$) response functions.
Second, we consider \emph{neural basis functions (MLP)}, where we fit a multi-layer perceptron with $K$ neurons in the last hidden layer to the observed data $\{(x_i, y_i)\}_{i \in N}$ and take $\psi_k(x)$ to be the activation of the $k$-th neuron in the last hidden layer.
We describe the details as well as an additional choice based on Gaussian process basis functions in Appendix~\ref{app:response}.

The choice of polynomials mainly illustrates a type of sensitivity analysis: we will contrast how bounds change when moving from a linear to quadratic, then quadratic to cubic and learned MLP representations.
Recall that finite linear combinations of basis functions can arbitrarily approximate infinite-dimensional function spaces.
The practitioner chooses its plausible complexity and pays the corresponding price in terms of looseness of bounds.
For instance, we can add as many knot positions for a regression splines procedure as we want to get arbitrarily close to nonparametric function spaces.
There is no concern for overfitting, given that data plays a role only via the estimation of $p(x, y\given z)$ or of particular black-box expectations (Appendix \ref{app:p_yxz}).
We emphasize that \emph{having a class of algorithms that allows for controlling the complexity of the function space is an asset, not a liability}.
Knowledge of functional constraints is useful even in a non-causal setting \citep{gupta:20}.
The linear basis formulation can be as flexible as needed, while allowing for shape and smoothness assumptions that are more expressive than all-or-nothing assumptions about, say, missing edges or additivity.
In Appendix~\ref{app:discretization}, we discuss an alternative based on discretization of $X$  combined with the off-the-shelf use of \cite{balke1994counterfactual}.
We demonstrate how in several ways that is just a \emph{less} flexible family of response functions than the approach discussed here, although see Appendix~\ref{app:p_eta} for a discussion on making $p_\eta(\theta \given x, z)$ also more flexible than the implementation discussed here.

\section{Experimental Results}
\label{sec:results}

\def\fracwidth{0.33}
\def\spacer{\hspace{2.cm}}
\def\vsqueeze{-1.5mm}
\begin{figure}
  \centering
  \includegraphics[width=\textwidth]{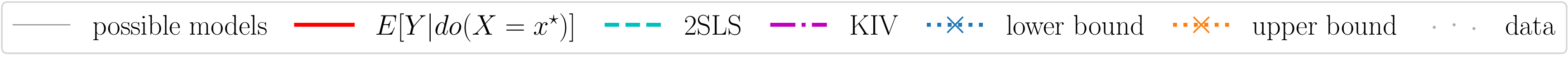}\\
  \textbf{linear response}\spacer \textbf{quadratic response}\spacer \textbf{MLP response}\\[1mm]
  \hrule\vspace{1mm}
  linear Gaussian setting with weak instrument and strong confounding $(\alpha\!=\!0.5, \beta\!=\!3)$\\
  \includegraphics[width=\fracwidth\textwidth]{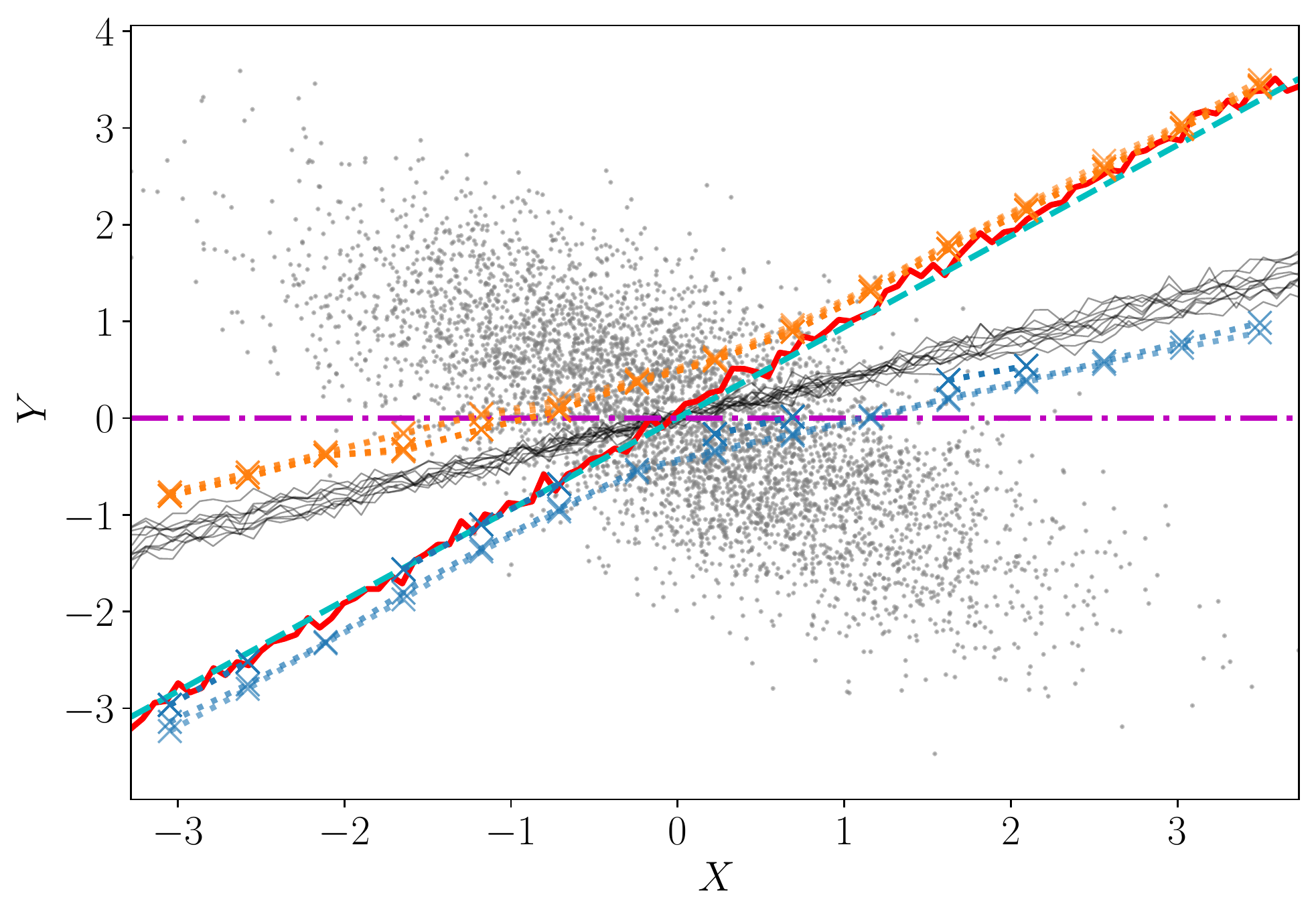}%
  \hfill
  \includegraphics[width=\fracwidth\textwidth]{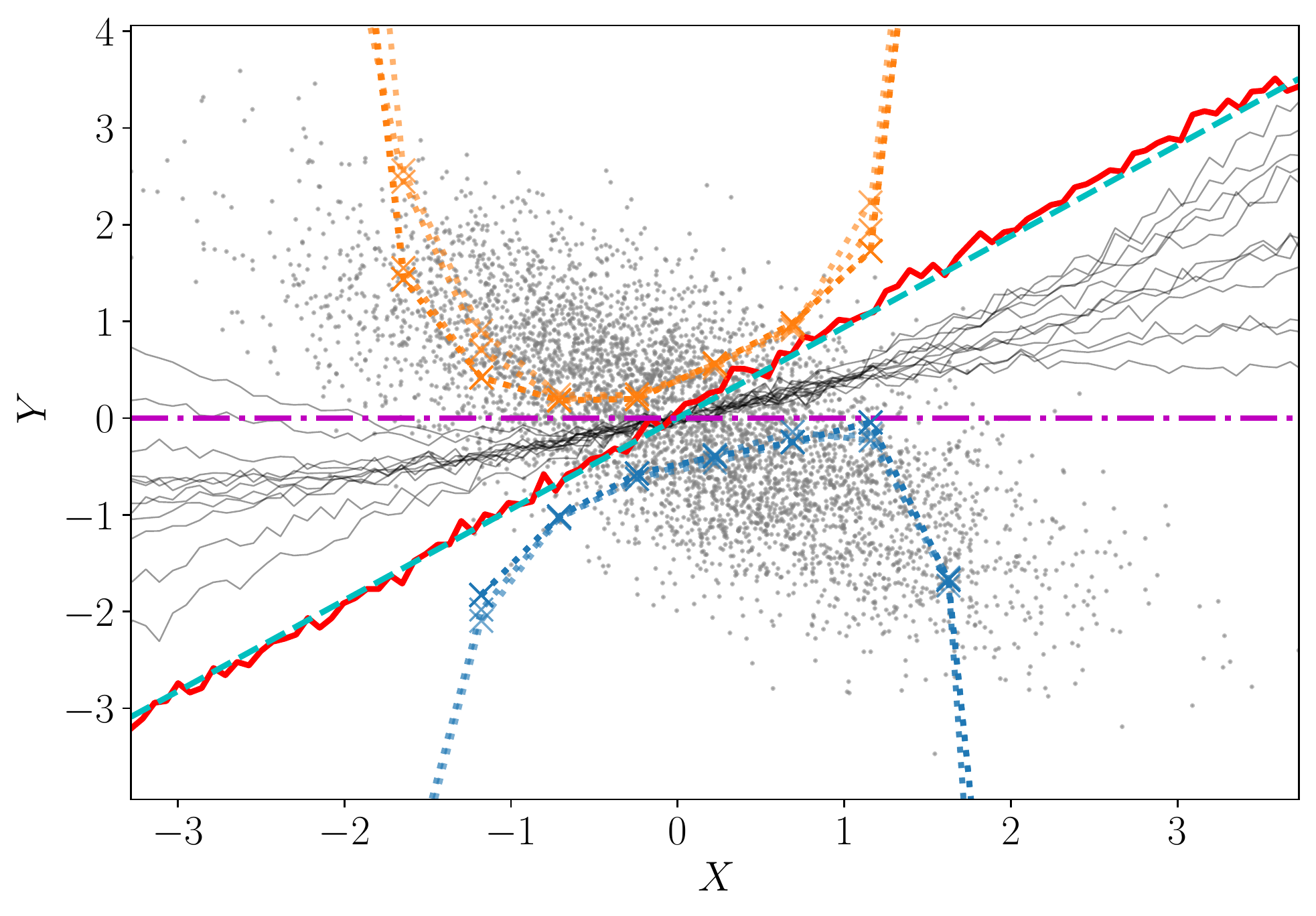}%
  \hfill
  \includegraphics[width=\fracwidth\textwidth]{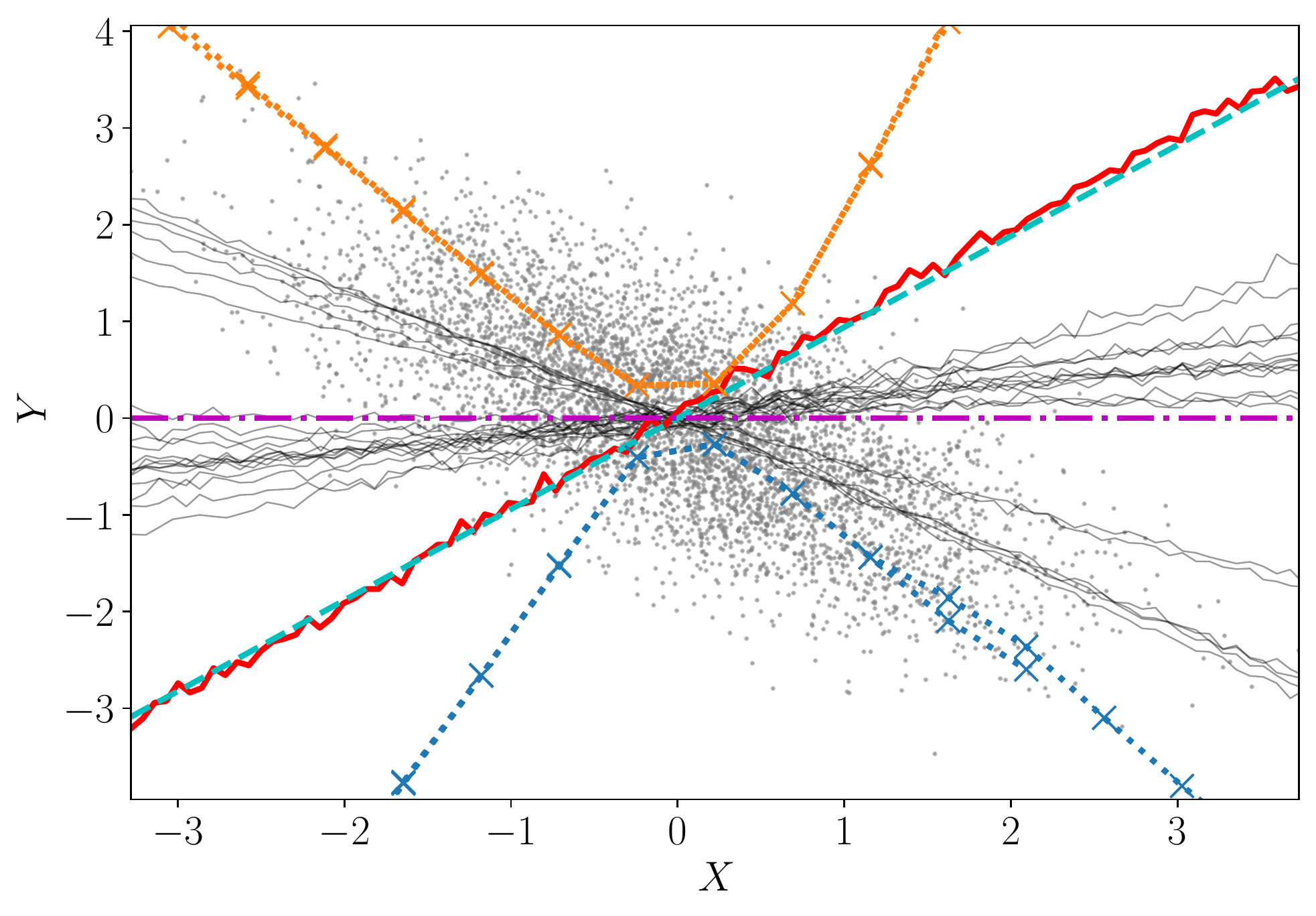}\\[\vsqueeze]
  linear Gaussian setting with strong instrument and weak confounding $(\alpha\!=\!3, \beta\!=\!0.5)$\\
  \includegraphics[width=\fracwidth\textwidth]{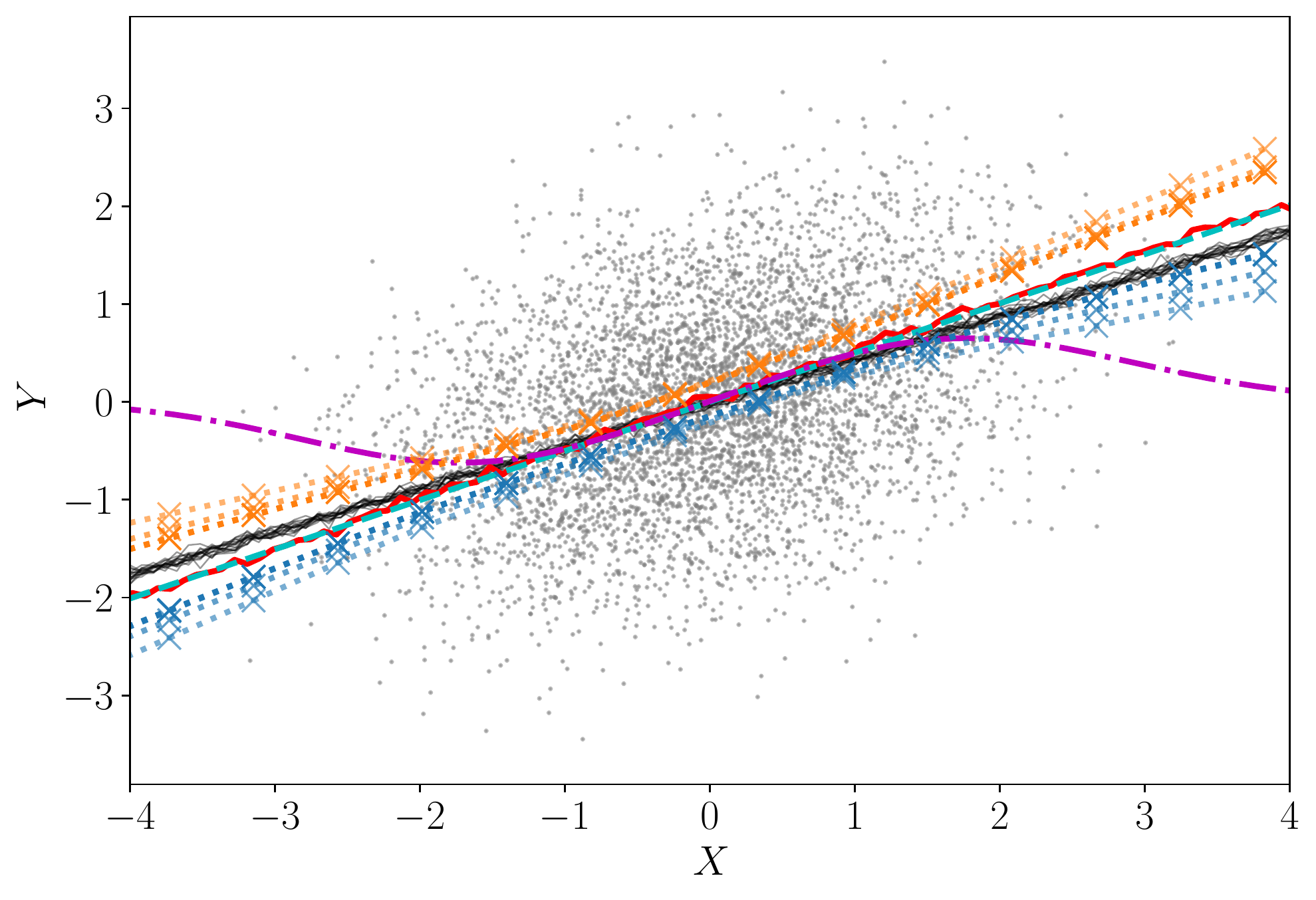}%
  \hfill
  \includegraphics[width=\fracwidth\textwidth]{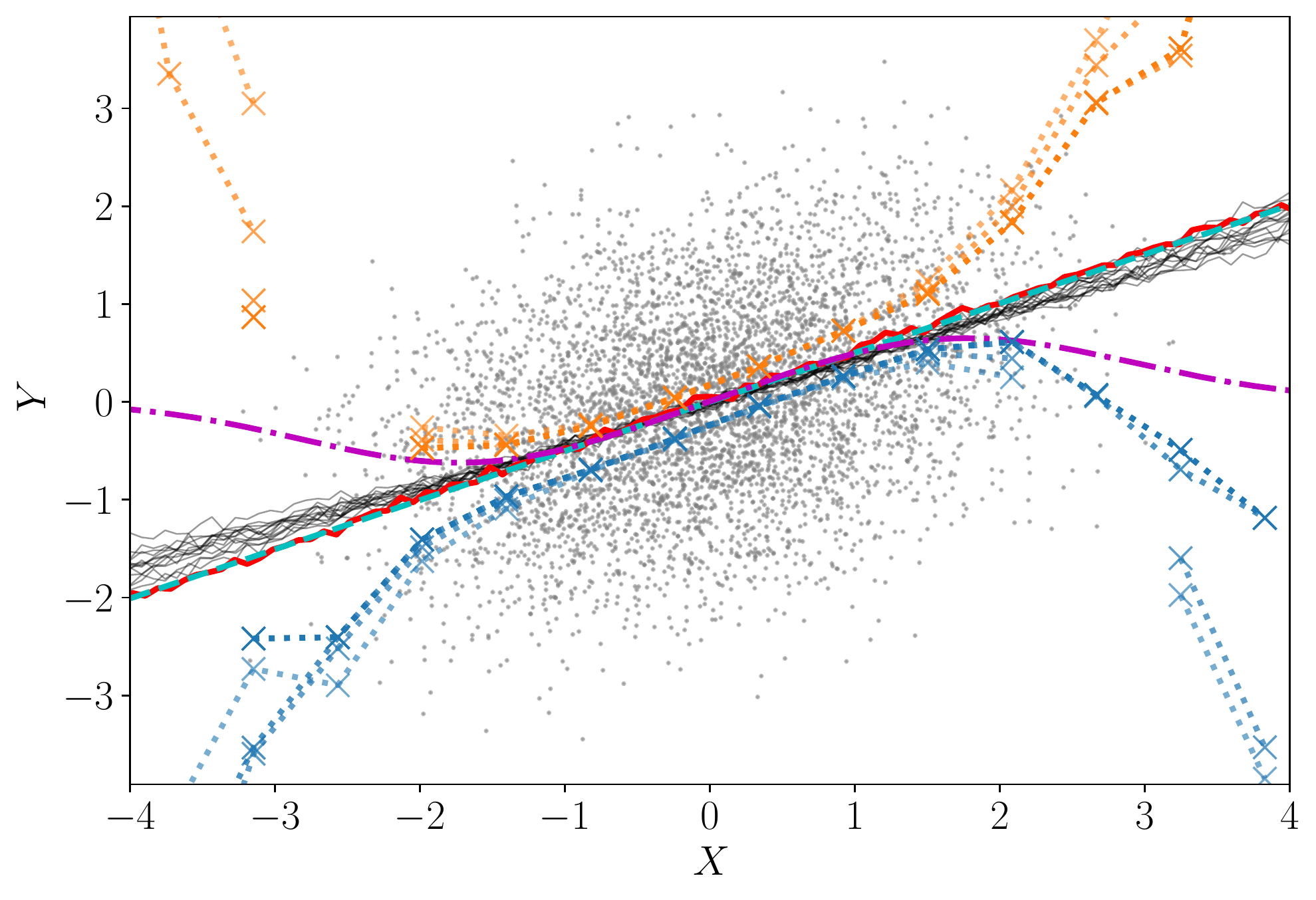}%
  \hfill
  \includegraphics[width=\fracwidth\textwidth]{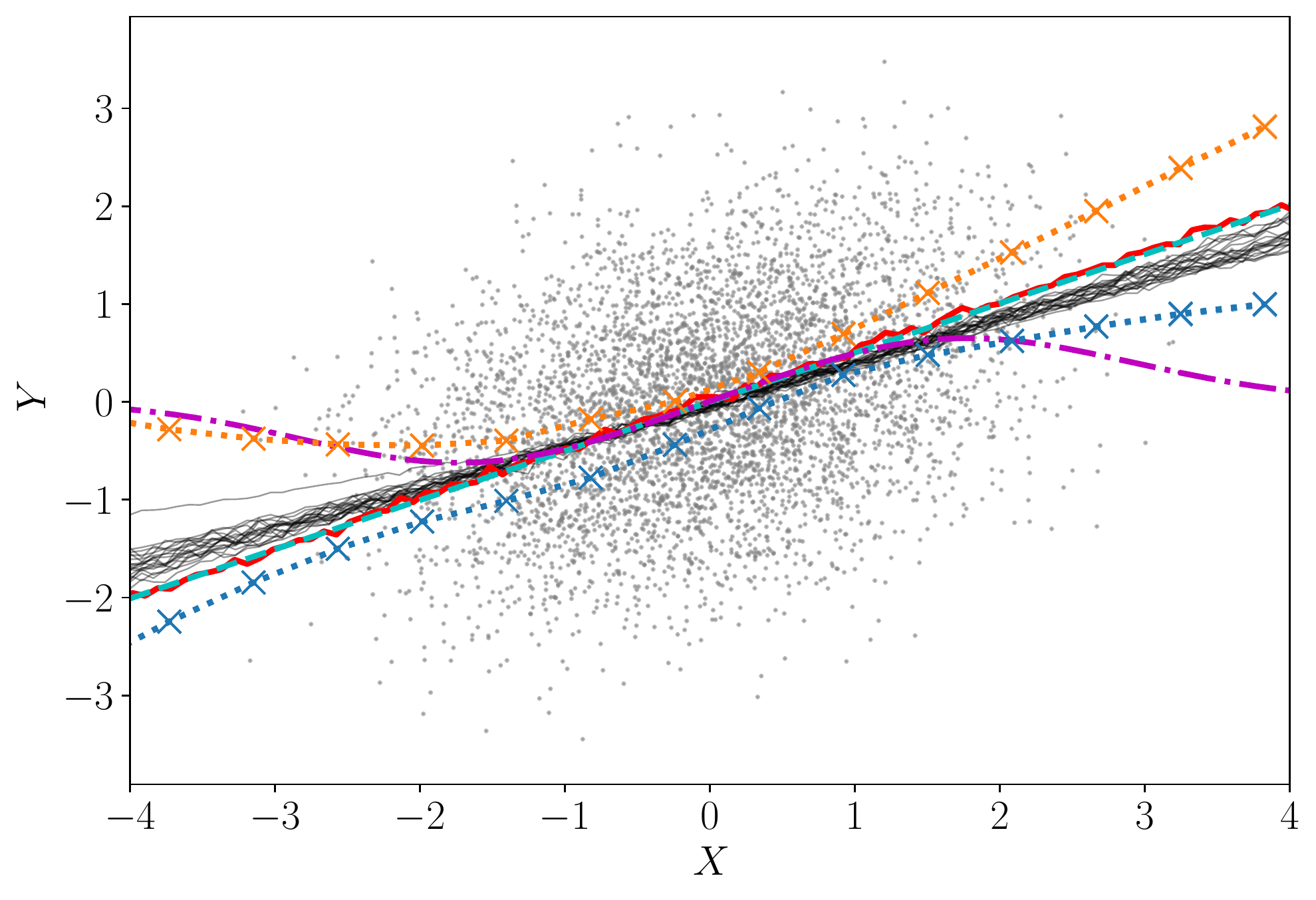}\\[\vsqueeze]
  non-additive, non-linear setting with weak instrument and strong confounding  $(\alpha\!=\!0.5, \beta\!=\!3)$\\
  \hfill
  \includegraphics[width=\fracwidth\textwidth]{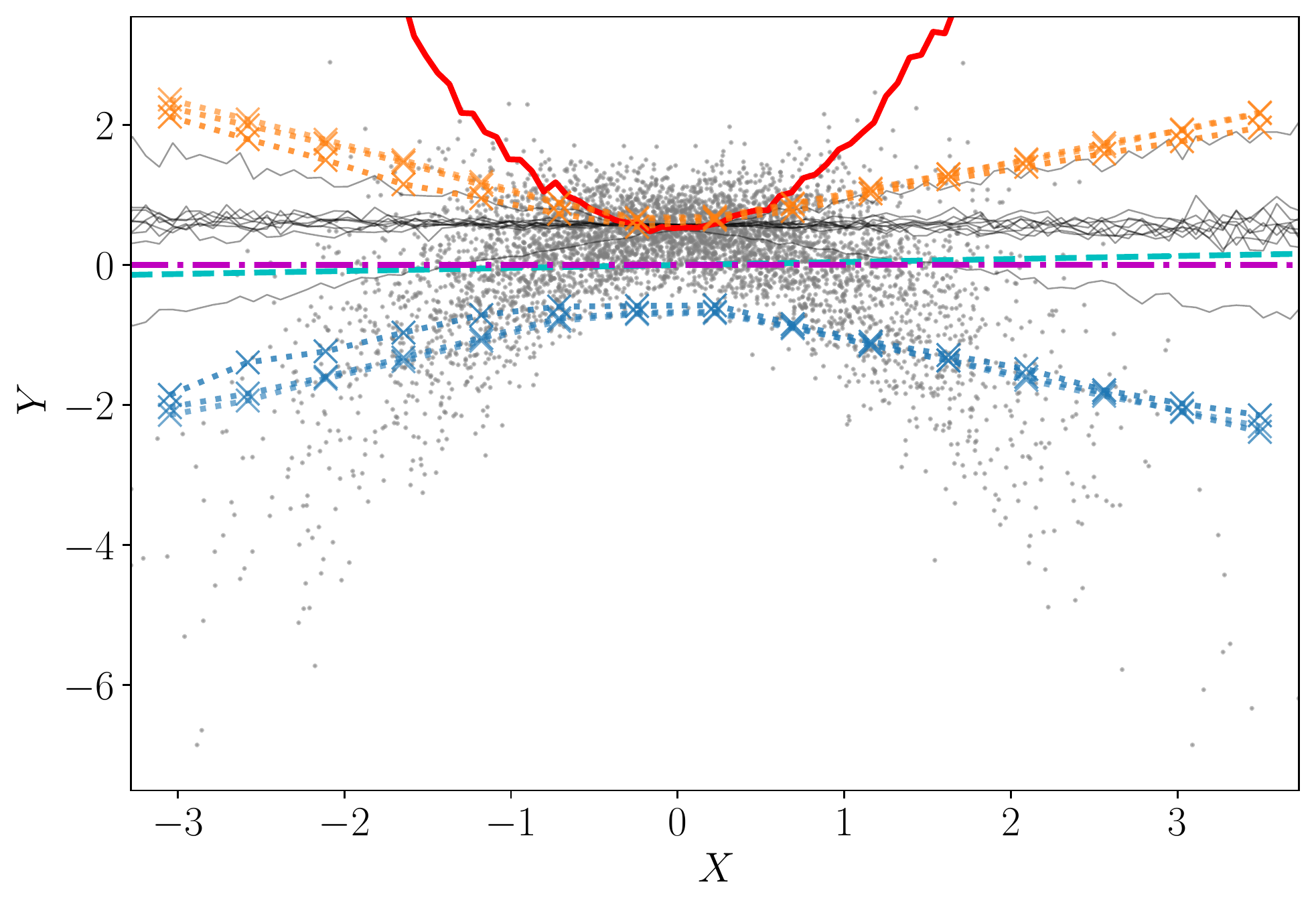}%
  \includegraphics[width=\fracwidth\textwidth]{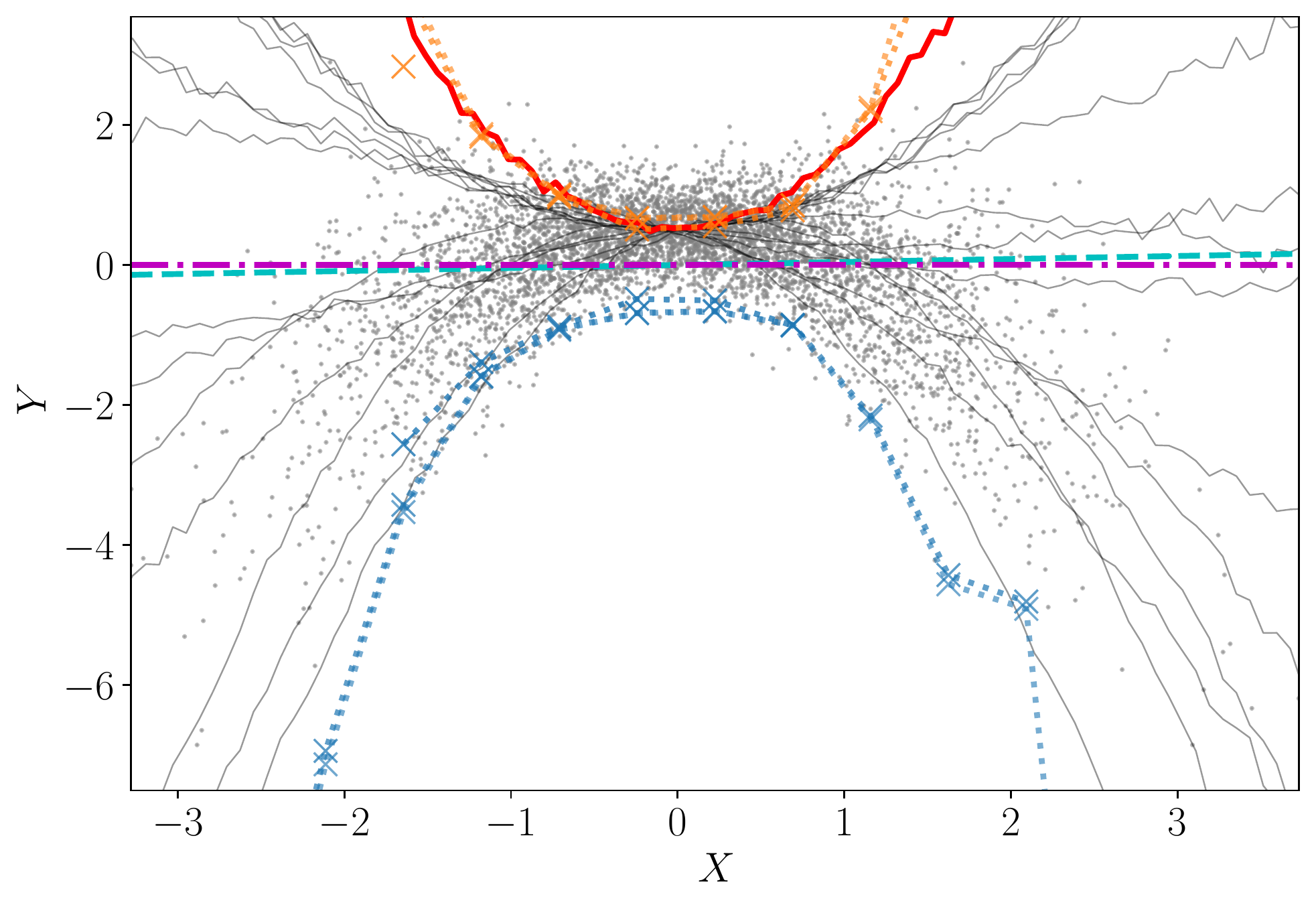}%
  \includegraphics[width=\fracwidth\textwidth]{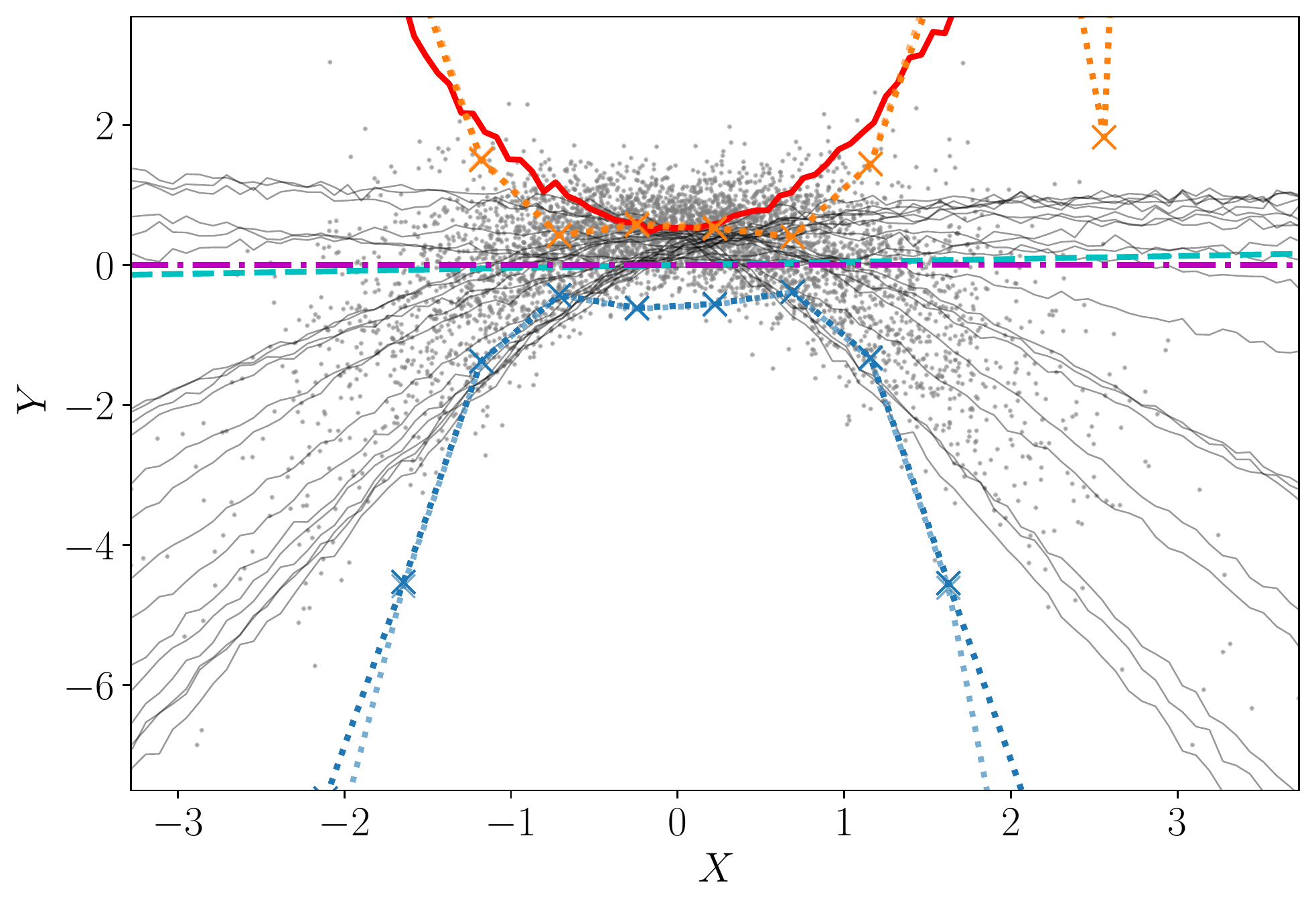}\\[\vsqueeze]
  non-additive, non-linear setting with strong instrument and weak confounding  $(\alpha\!=\!3, \beta\!=\!0.5)$\\
  \hfill
  \includegraphics[width=\fracwidth\textwidth]{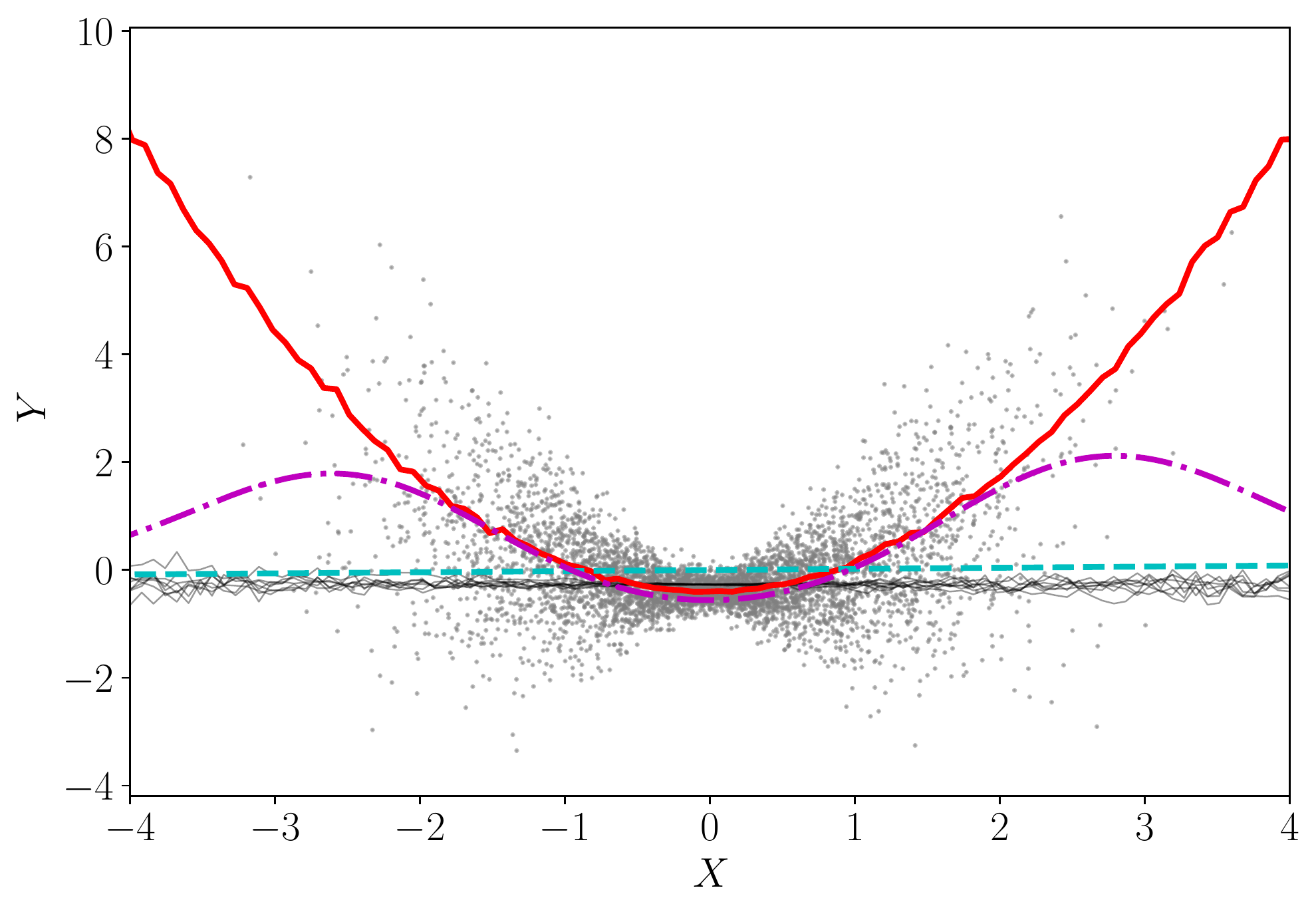}%
  \includegraphics[width=\fracwidth\textwidth]{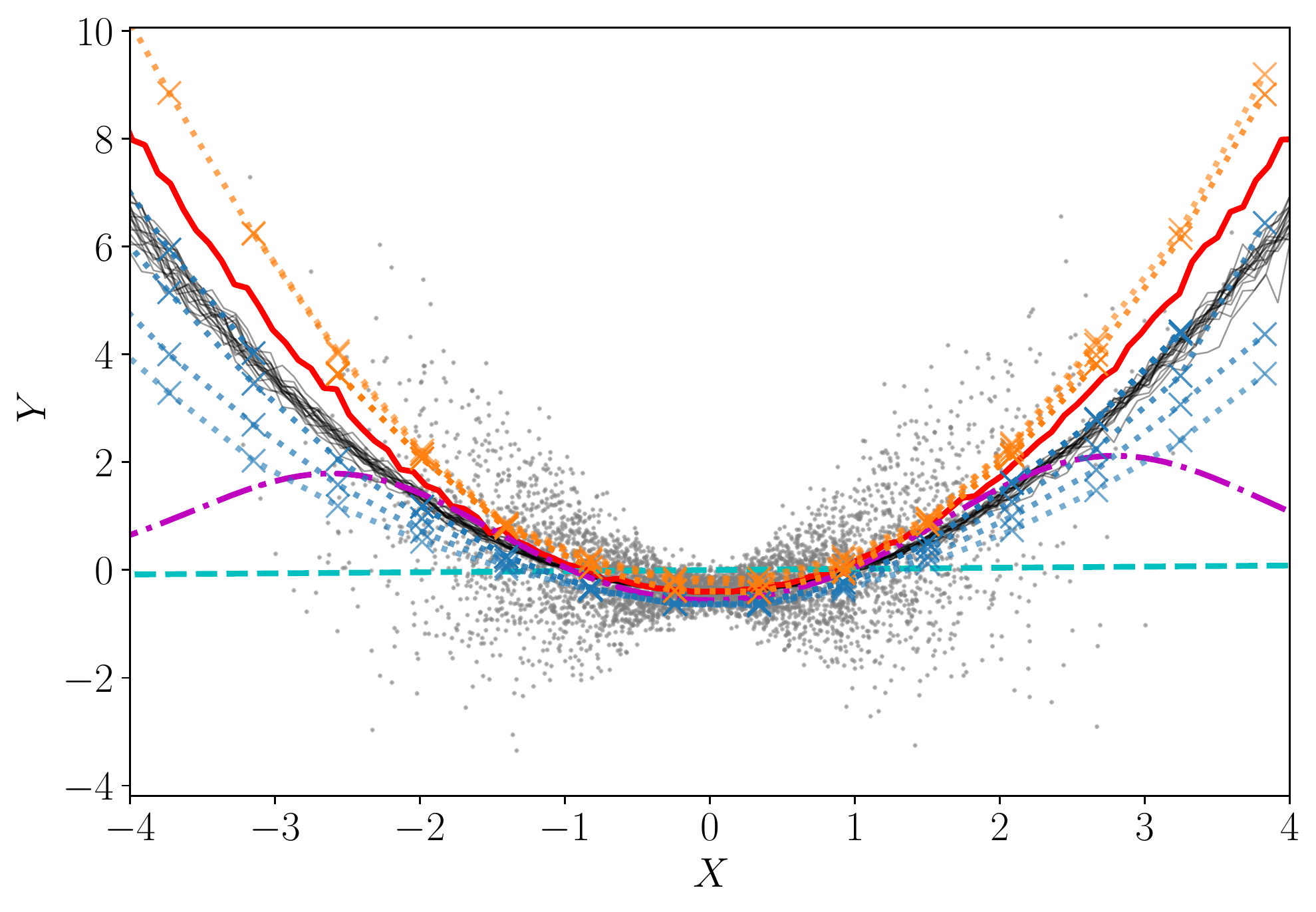}%
  \includegraphics[width=\fracwidth\textwidth]{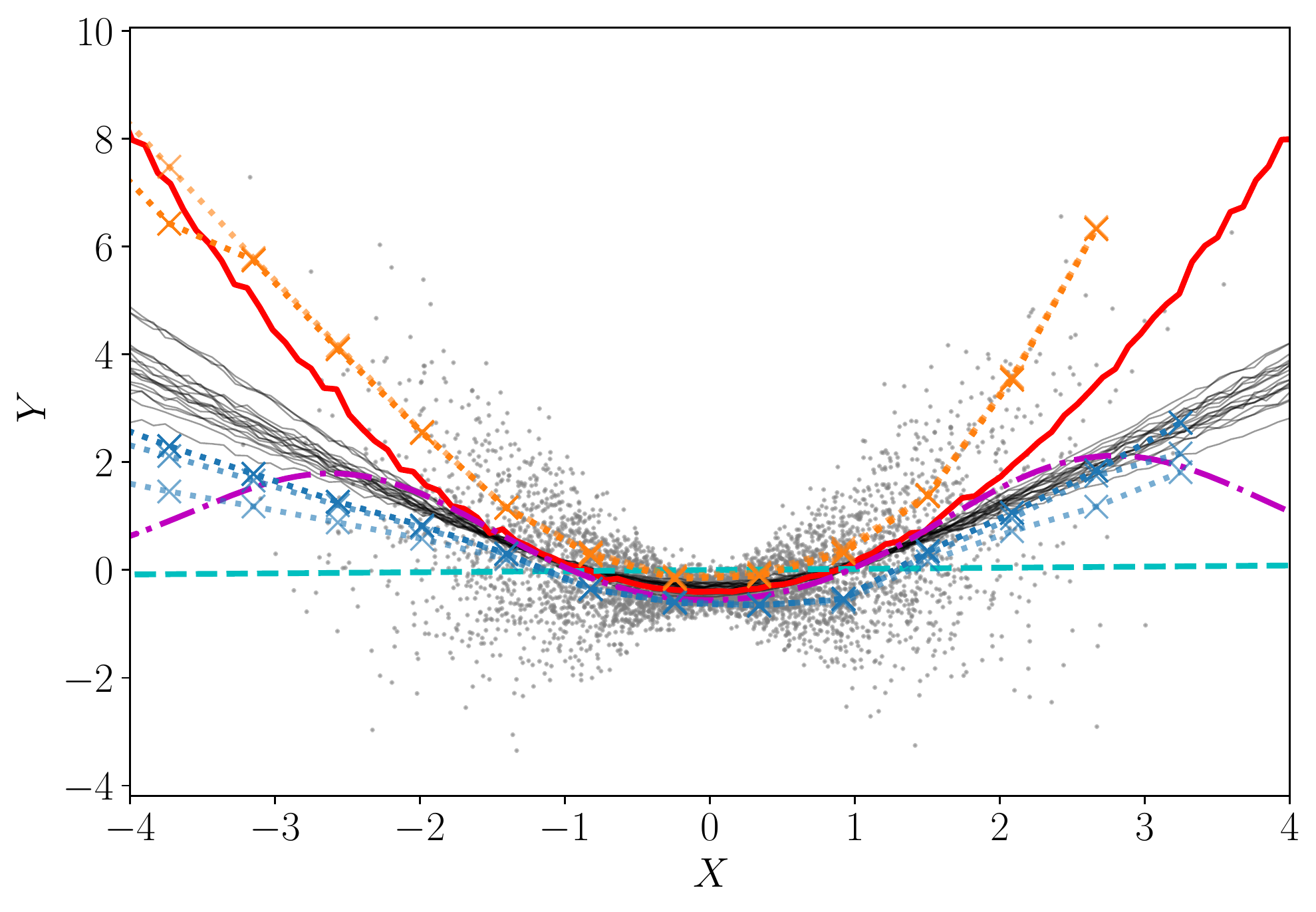}%
  \caption{Results for synthetic datasets (linear Gaussian and non-linear, non-additive) for a weak and strong instrument respectively.
  Columns correspond to different response function families.}
  \label{fig:bounds_synth}
\vspace{-0.20in}
\end{figure}

We evaluate our method on a variety of synthetic and real datasets.
In all experiments, we report the results of two stage least squares (\textbf{2SLS} \tslsline{}) and kernel instrumental variable regression (\textbf{KIV} \kivline{}) \citep{singh2019kernel}.
Note that both methods assume additive noise and provide point estimates for expected outcomes under a given treatment.
The KIV implementation by \citet{singh2019kernel} comes as an off-the-shelf method with internal heuristics for tuning hyperparameters.
For our method, we show \textbf{lower} (\lowerline{}) and \textbf{upper} (\upperline{}) bounds computed individually for multiple values of $x^{\star} \in \bR$.
The transparency of these lines indicates the tolerances $\abstol, \reltol$, where more transparency corresponds to larger tolerances.
Missing bounds at an $x^{\star}$ indicate that the constraints could not be satisfied in the optimization.
In the synthetic settings, we also show the \textbf{true causal effect} $\E[Y \given do(X \!=\! x^{\star})]$ (\trueline{}).

Finally, we highlight that there are multiple possible causal effects compatible with the data (which our method aims to bound).
To do so, we fit a latent variable model of the form shown in Figure~\ref{fig:setup}(a) to the data, with $U \given Z, X, Y \sim \cN(\mu(Z, X, Y), \sigma^2(Z, X, Y))$ where $\mu, \sigma^2$ as well as $\E[X \given Z, U]$ are parameterized by neural networks.
We ensure that the form of $\E[Y \given X, U]$ matches our assumptions on the function form of the response family (i.e., either polynomials of fixed degree in $X$, or neural networks).
We then optimize the evidence lower bound following standard techniques \citep{kingma2013auto}, see Appendix~\ref{app:latent}.
We fit multiple models with different random initializations and compute \textbf{the implied causal effect} of $X$ on $Y$ for each one, shown as multiple thin gray lines (\modelline{}).
We report results for additional datasets as well as how our method performs in the small data regime in Appendix~\ref{app:results}.
All experiments use a single set of hyperparameters, see Appendix~\ref{app:hyperparams}.

\xhdr{Linear Gaussian case}
First, we test our method in a synthetic linear Gaussian scenario, where instrument, confounder, and noises $Z, C, e_X, e_Y$ are independent standard Gaussian variables.
We consider two settings of the form $X = g(Z, C, e_X) := \alpha \, Z + \beta \, C + e_X$ and $Y = f(X, C, e_Y) := X - 6 \, C + e_Y$, with $\alpha, \beta \in \{(0.5, 3), (3, 0.5)\}$.
The two settings of coefficients $\alpha, \beta$ describe a weak instrument with strong confounding and a strong instrument with weak confounding respectively.
The first two rows of Figure~\ref{fig:bounds_synth} show our bounds in these settings for linear, quadratic and MLP response functions.
Because these scenarios satisfy all theoretical assumptions of 2SLS and KIV, 2SLS (\tslsline{}) reliably recovers the true causal effect, which is simply $\E[Y \given do(X=x^{\star})] = x^{\star}$.
For a weak instrument, KIV (\kivline{}) fails by reverting to its prior mean $0$ everywhere, whereas it matches the true effect in data rich regions in the second setting with weak confounding.\footnote{We provide more details on this failure mode of KIV in Appendix~\ref{app:optimistickiv}.}

We observe that the true causal effect (\trueline{}) is always within our bounds (\lowerline{}, \upperline{}).
Moreover, our bounds also contain most of the ``other possible models'' that could explain the data (\modelline{}), showing that they are highly informative, without being more confident than warranted.
As expected, our bounds get looser as we increase the flexibility of the response functions (linear, quadratic, MLP from columns~1-3).
In particular, allowing for flexible MLP responses (column~3), our bounds are rightfully loose for strong confounding.
As confounding weakens and the instrument strengthens (in the second row) the gap between our bounds gets narrower.

\xhdr{Non-additive, non-linear case}
Our next synthetic setting is non-linear and violates the additivity assumption.
Again, the treatment is given by $X = \alpha \, Z + \beta \, C + e_X$ with the same set of coefficients $\alpha, \beta$ as for the linear setting.
The outcome is non-linear and non-additive $Y = 0.3 \, X^2 - 1.5 \, X \, C + e_Y$ with a true effect of $\E[Y \given do(X=x^{\star})] = 0.3\, (x^{\star})^2$.
The bottom two rows of Figure~\ref{fig:bounds_synth} show our results for this setting.
Since additivity is violated (due to the $X \, C$-term) and the effect is non-linear, 2SLS fails.
Without additivity, KIV also fails for strong confounding, but captures the true effect well in data rich regions when the instrument is strong and confounding is weak.
The strongly confounded case (row~3) highlights the effect of the choice of response functions.
Wrongly assuming linear response functions, our bounds rule out the true effect (row~3, column~3).
However, they capture the implied causal effects from possible compatible linear models.
As we allow for more flexible response functions capable of describing the true effect, our bounds are extremely conservative (row~3, columns~2 \& 3) as they should be, indicated by the effects from other compatible models.
In the strong instrument, weak confounding case (row~4), our bounds become narrower to the point of essentially identifying the true effect for adequate response functions (column~2).
Here, linear response functions cannot explain the data anymore, indicated by missing bounds (row~4, column~1).

\def\fracwidth{0.33}
\def\spacer{\hspace{2.2cm}}
\begin{figure}
  \centering
  linear response\spacer quadratic response \spacer MLP response\\
  \hfill
  \includegraphics[width=\fracwidth\textwidth]{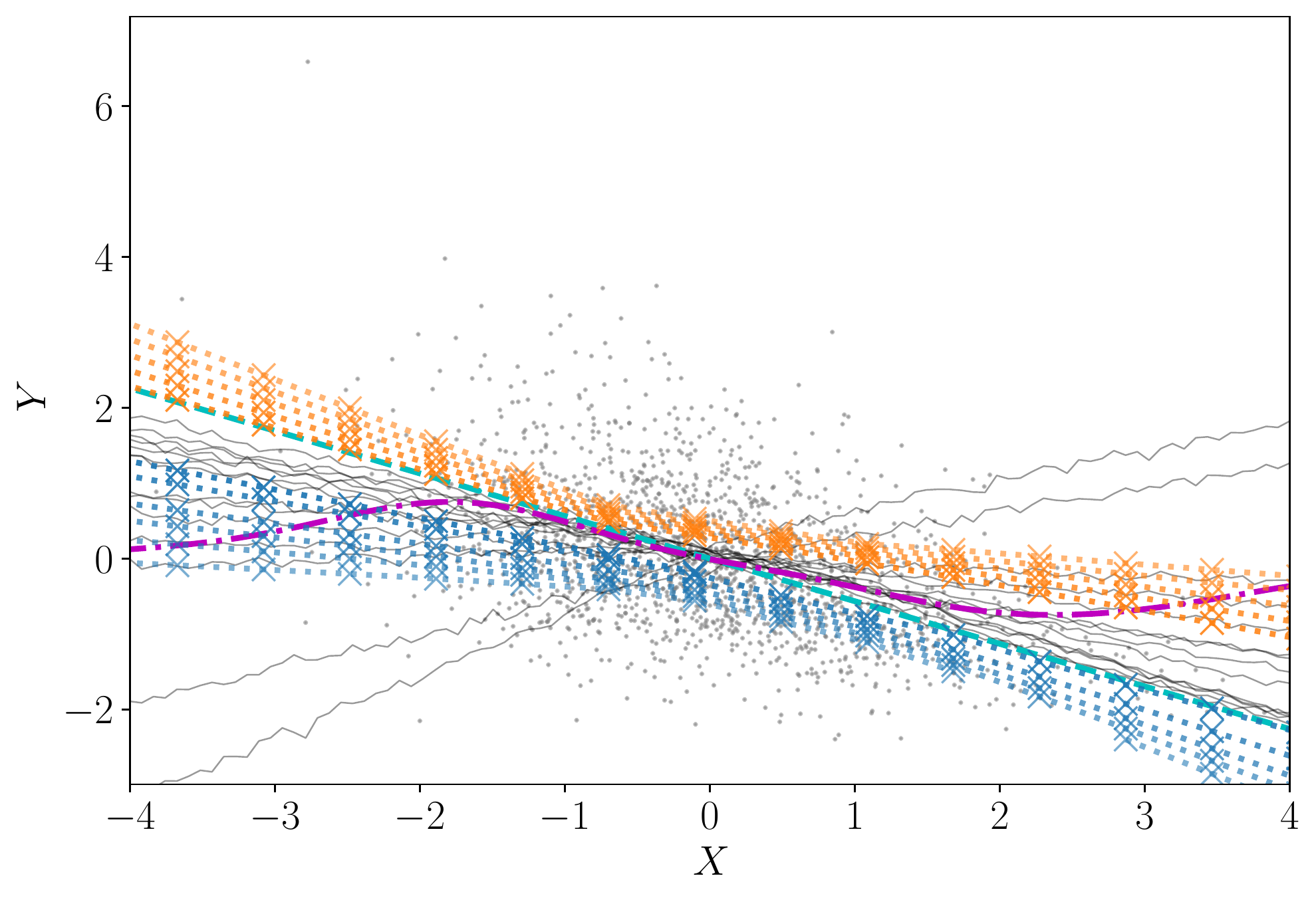}%
  \hfill
  \includegraphics[width=\fracwidth\textwidth]{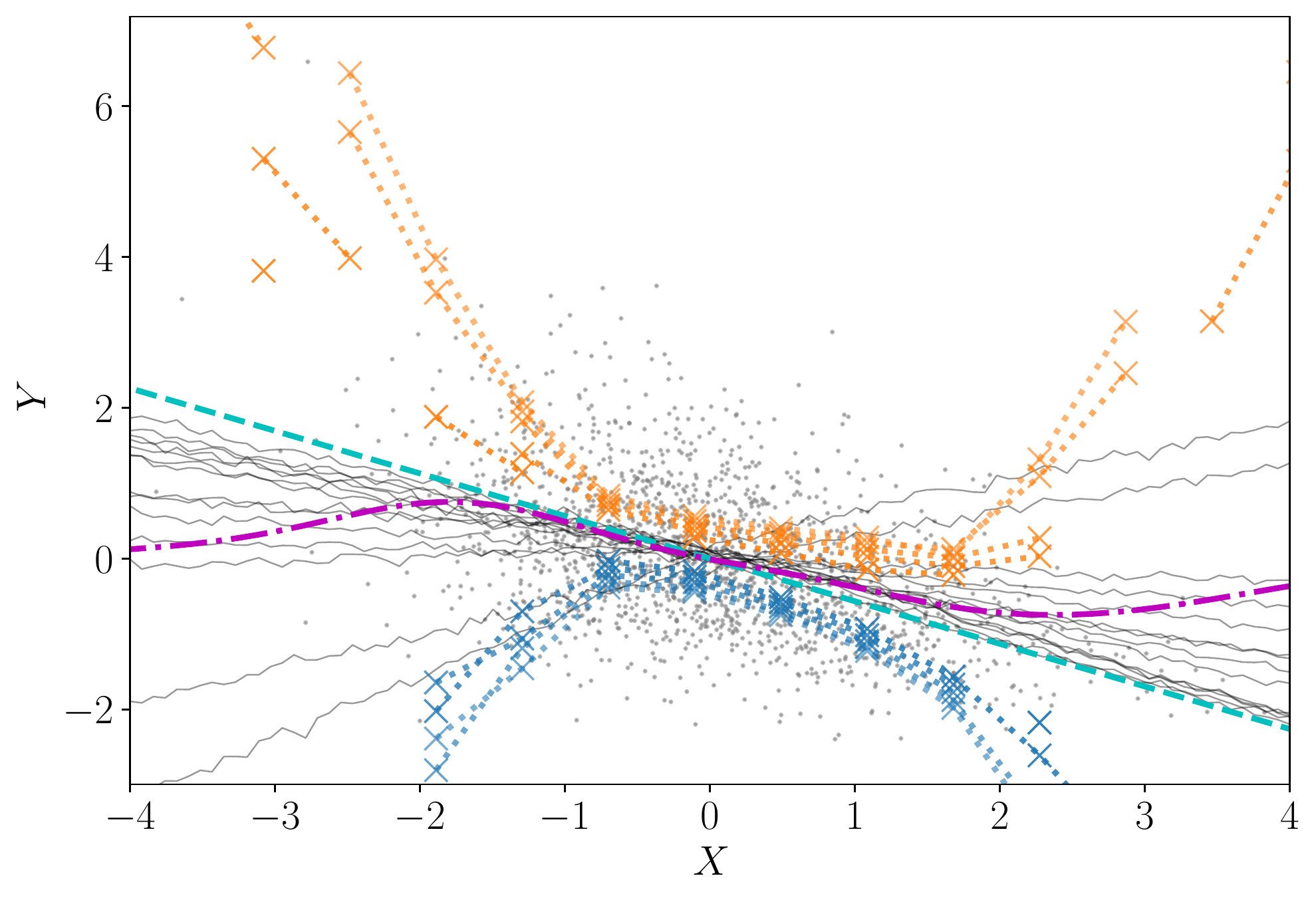}%
  \hfill
  \includegraphics[width=\fracwidth\textwidth]{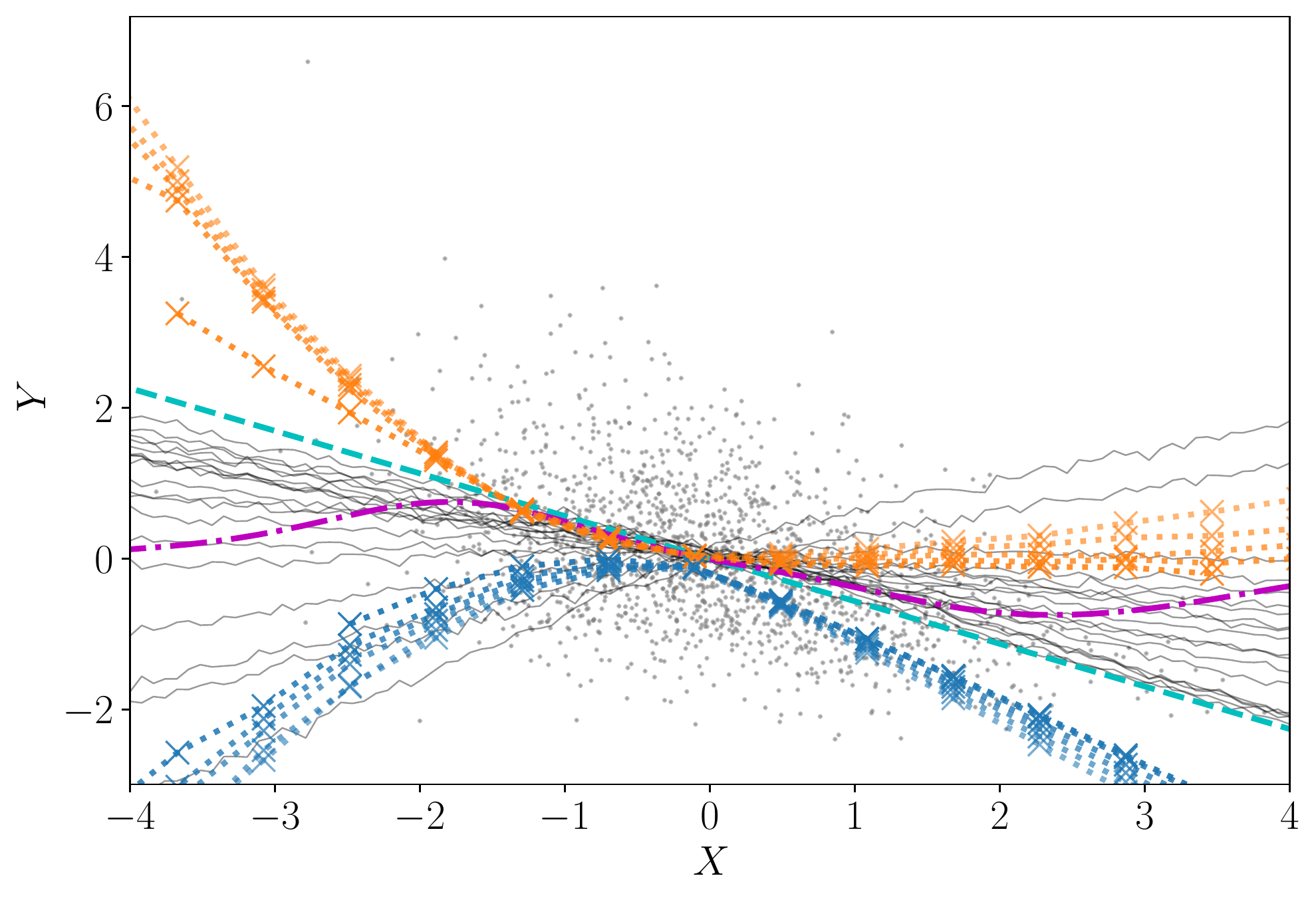}%
  \caption{Results on the expenditure dataset for different response function families.}
  \label{fig:bounds_expenditure}
\end{figure}

\xhdr{Expenditure data}
We now turn to a real dataset from a 1995/96 survey on family expenditure in the UK \citep{expenditure2000}.
This dataset has been used by \citet{gunsilius2019bounds} and previously \citep{blundell2007semi,imbens2009identification} for 1994/95 data.
The outcome of interest is the share of expenditure on food.
The treatment is the log of the total expenditure and the instrument is gross earnings of the head of the household.
All three variables are continuous, relations cannot be expected to be linear, and we cannot exclude unobserved confounding \citep{gunsilius2019bounds}, making this a good test case for our method.
We describe the data in more detail in Appendix~\ref{app:data}.
Figure~\ref{fig:bounds_expenditure} shows that our bounds provide useful information about both the sign and magnitude of the causal effect and gracefully capture the increasing uncertainty as we allow for more flexible response functions.
Moreover, they include most of the possible effects from latent variable models indicating that they are not overly restrictive.
The few curves that escaped our bounds correspond to situations where the latent variable model fit was suboptimal in terms of local likelihood and hence may be an artifact of the latent variable model training procedure.

\section{Conclusion}
\label{sec:conclusion}
We have proposed a class of algorithms for computing bounds on causal effects by exploiting modern optimization machinery. 
While this addresses an important source of uncertainty in causal inference --- partial identifiability as opposed to full identifiability --- there is also statistical uncertainty: confidence or credible intervals for the \emph{bounds} themselves \citep{imbens:04}.
Clearly this is an important matter to be addressed in future work, and the black-box approach of \cite{silva:16} provides some directions for credible intervals.
There are also considerations about the parameterization of $p_\eta(\theta \given x, z)$ and how possible pre-treatment covariates can be non-trivially used in the model.
We defer these considerations to Appendix~\ref{app:p_eta}.
Other parameterizations of the IV model, such as the one by \cite{zhang:20} can lead to alternative algorithms and ways of expressing assumptions.

One could also use the same ideas to test whether an IV model is valid, another common use of latent variable causal models \citep[e.g.,][]{wolfe2019inflation}.
We assumed that the model was correct.
Model falsification can still be done, which will happen when the optimization fails to find a solution \citep{silva:16}, and observed in some of the experiments reported.
Specializing methods for testing models instead of deriving bounds is an interesting direction for future work.

Finally, we foresee our ideas as ways of liberating causal modeling to accommodate ``softer,'' more general constraints than conditional independence statements.
For instance, as described by \cite{silva:16}, there is no need to assume any sparsity in a causal DAG, as long as we know that some edges are ``weak'' (in a technical sense) so that, e.g., edge $Z \rightarrow Y$ is allowed, but its influence on $Y$ is not arbitrary.
How to do that in a computationally feasible way remains a challenge, but the possibility of complementing causal inference based on sparse DAGs, such as the do-calculus of \cite{pearl2009causality}, with the sledgehammer of modern continuous optimization, is an attractive prospect.

\section*{Broader Impact}
Cause effect estimation is crucial in many areas where data-driven decisions may be desirable such as healthcare, governance or economics.
These settings commonly share the characteristic that experimentation with randomized actions is unethical, infeasible or simply impossible.
One of the promises of causal inference is to provide useful insights into the consequences of hypothetical actions based on observational data.
However, causal inference is inherently based on assumptions, which are often untestable.
Even a slight violation of the assumptions may lead to drastically different conclusions, potentially changing the desired course of action.
Especially in high-stakes scenarios, it is thus indispensable to thoroughly challenge these assumptions.

This work offers a technique to formalize such a challenge of standard assumptions in continuous IV models.
It can thus help inform highly-influential decisions.
One important characteristic of our method is that while it can provide informative bounds under certain assumptions on the functional form of effects, the bounds will widen as less prior information supporting such assumptions is available.
We can view this as a way of deferring judgment until stricter assumptions have been assessed and verified.

Since our algorithms are causal inference methods, they requires assumptions too.
Therefore, our method also requires a careful assessment of these assumptions by domain-experts and practitioners.
In addition, as we are optimizing a non-convex problem with local methods, we have no theoretical guarantee of correctness of our bounds.
Hence, if wrong assumptions for our model are accepted prematurely, or our optimization strategy fails to find global optima, our method may wrongly inform decisions.
If these are high-stakes decisions, then wrong decisions can have significant negative consequences (e.g., a decision not to treat a patient that should be treated).
If the data that this model is trained on is biased against certain groups (e.g., different sexes, races, genders) this model will replicate those biases.
We believe a fruitful approach towards making our model more sensitive to uncertainties due to structurally-biased, unrepresentative data, is to learn how to derive, then inflate (to account for bias)  uncertainty estimates for our bounds.

\begin{ack}
We thank Florian Gunsilius for useful discussions, providing code for his method and explaining how to prepare the Family Expenditure Survey dataset. We are grateful to Robin Evans, Arthur Gretton, Jiri Hron, Paul Rubenstein, and Rahul Singh for useful discussions and feedback.
MK and RS acknowledge support from the The Alan Turing Institute under EPSRC grant EP/N510129/1.
This work was partially done while RS was on a sabbatical at the Department of Statistics, University of Oxford.
\end{ack}

\bibliography{refs}
\bibliographystyle{icml2020}

\clearpage
\appendix
\section{Gunsilius's Algorithm}
\label{app:gunsilius}

\cite{gunsilius2019bounds} provides a theoretical framework for minimal conditions for a continuous IV model to imply non-trivial bounds (that is, bounds tighter that what can be obtained by just assuming that the density function $p(x, y \given z)$ exists).
That work also introduces two variations of an algorithm for fitting bounds.

The basic version consists of first sampling $l$ response functions $f_{R_x}(\cdot)$ and $f_{R_y}(\cdot)$ from a distribution over functions -- in the experiments described, a Gaussian process evaluated on a grid in the respective spaces.
The final distribution is reweighted combination of the pre-sampled $l$ response functions with weights $\mu$ playing the role of the decision variables to be optimized.
Hence, by construction, the space of distributions in the response function space is absolutely continuous with respect to the pre-defined Gaussian process.
The constraints are defined by approximating an estimate of the bivariate CDF $F(x, y \given z)$ on a grid of values, which are approximately constrained to match the model implied CDF in a $L_2$ sense.
Large deviance bounds are then used to show the (intuitive) result that this approximation is a probably approximately correct formulation of the original optimization problem.

One issue with this algorithm is that $l$ may be required to be large as it is a non-adaptive Monte Carlo approximation in a high dimensional space.
A variant is described where, every time a solution for $\mu$ is found, response function samples with low corresponding values of $\mu$ are replaced (again, from the given and non-adaptive Gaussian process).
Although this now has the advantage of adapting the Monte Carlo samples to the problem, this has convergence problems that may be severe and not easy to diagnose.

In contrast, we formulate our adaptation of $\eta$ as a continuous optimization problem with an estimate of the gradient that has empirically reasonable stability, as expected from the related work in the machine learning literature for gradient estimation.
We also parameterize the distribution so that the only constraint that we need to enforce concerns the univariate density $p(y \given z)$ (or $p(y \given x, z)$, in the variation discussed in Appendix \ref{app:p_yxz}, which in principle requires no density estimation).
Like the algorithm given by Gunsilius, the space of functions is a linear combination of a fixed dictionary of basis functions with a Gaussian distribution on the parameters, although we do not make use of the discrete mixture reweighting on the Monte Carlo samples, which introduces instability in \citep{gunsilius2019bounds} despite its good theoretical properties.
Our formulation, like the one in \citep{gunsilius2019bounds}, can in principle make use of more a flexible distribution such as a mixture of Gaussian copulas at the cost of more computation, as discussed in Appendix~\ref{app:p_eta}.
An important piece of future work is to thoroughly assess how stable a mixture of Gaussians version our algorithm is in practice.

The proposed implementation of Gunsilius' algorithm computes $F_{Y \given do(x^{\star}_0)}(y^{\star}) - F_{Y \given do(x^{\star}_1)}(y^{\star})$, i.e., the difference in effects at two different treatment levels $x^{\star}_0$ and $x^{\star}_1$ for individuals within a fixed quantile $y^{\star} \in [0, 1]$ of the outcome variable.
For example, in the expenditure dataset (see Section~\ref{app:data}), the setting $x^{\star}_0 = 0.75, x^{\star}_1 = 0.25, y^{\star}= 0.25$ would look at how much people, who spend a lot overall ($x^{\star}= 0.75$) and spend comparably little on food (up to 25\%), would spend on food relatively to overall expenditure, if they spent much less overall ($x^{\star}_1 = 0.25$).
The main tuning parameter in the proposed algorithm is the penalization parameter $\lambda$, which corresponds to the tightness of the constraint.
In the proposed implementation, this parameter is fixed throughout the optimization and must be chosen manually.
In Figure~\ref{fig:gunsilius}, we show the results of Gunsilius's algorithm for three different levels of $y^{\star}$ on the expenditure dataset.
Small values of $\lambda$ result in uninformatively loose bounds and do not always seem to converge (e.g., for $y^{\star} = 0.75$).
As we increase $\lambda$, which corresponds to stronger enforcement of the constraint, the bounds get narrower.
However, even after a long burn-in period, we still encounter substantial ``instantaneous jumps'' as well as longer-term drifts in the bounds, which may change the qualitative conclusions (for example in the $y^{\star} = 0.75$ setting).
Note that this algorithm works on the empirical CDFs of all variables, i.e., they are all scaled to lie within $[0,1]$.

Moreover, even after laboriously improving the performance of the algorithm using acceleration via JAX \citep{jax2018github} and parallelized solving of the quadratic programs with CVXPY \citep{diamond2016cvxpy}, producing an upper and lower bound for a single setting of $x^{\star}_0, x^{\star}_1, y^{\star}, \lambda$ with Gunsilius's algorithm took longer (about 30 minutes on a quad-core Intel Core i7) than a full set of upper and lower bounds at 15 different $x^{\star}$ values with our algorithm (about 20 minutes on the same hardware).

\def\fracwidth{0.325\textwidth}
\def\spacer{\hspace{3cm}}
\begin{figure}
  \centering
  \includegraphics[width=\textwidth]{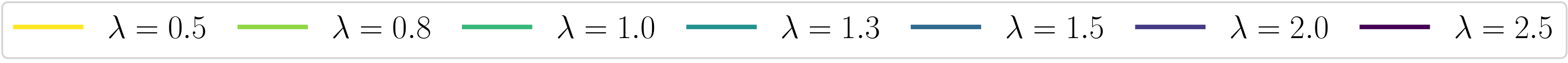}\\
  $y^{\star} = 0.25$ \spacer $y^{\star} = 0.5$ \spacer $y^{\star} = 0.75$\\
  \includegraphics[width=\fracwidth]{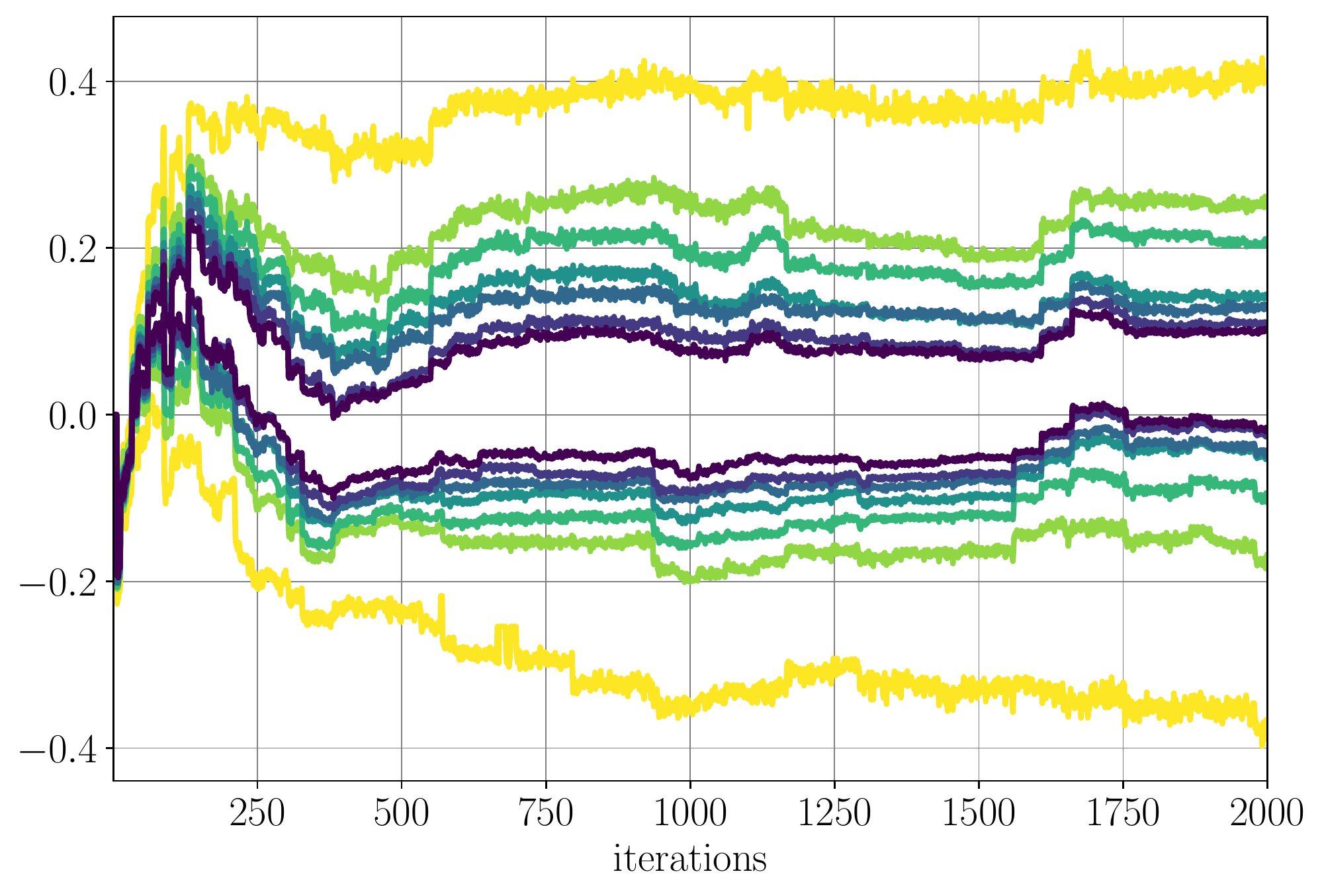}%
  \hfill
  \includegraphics[width=\fracwidth]{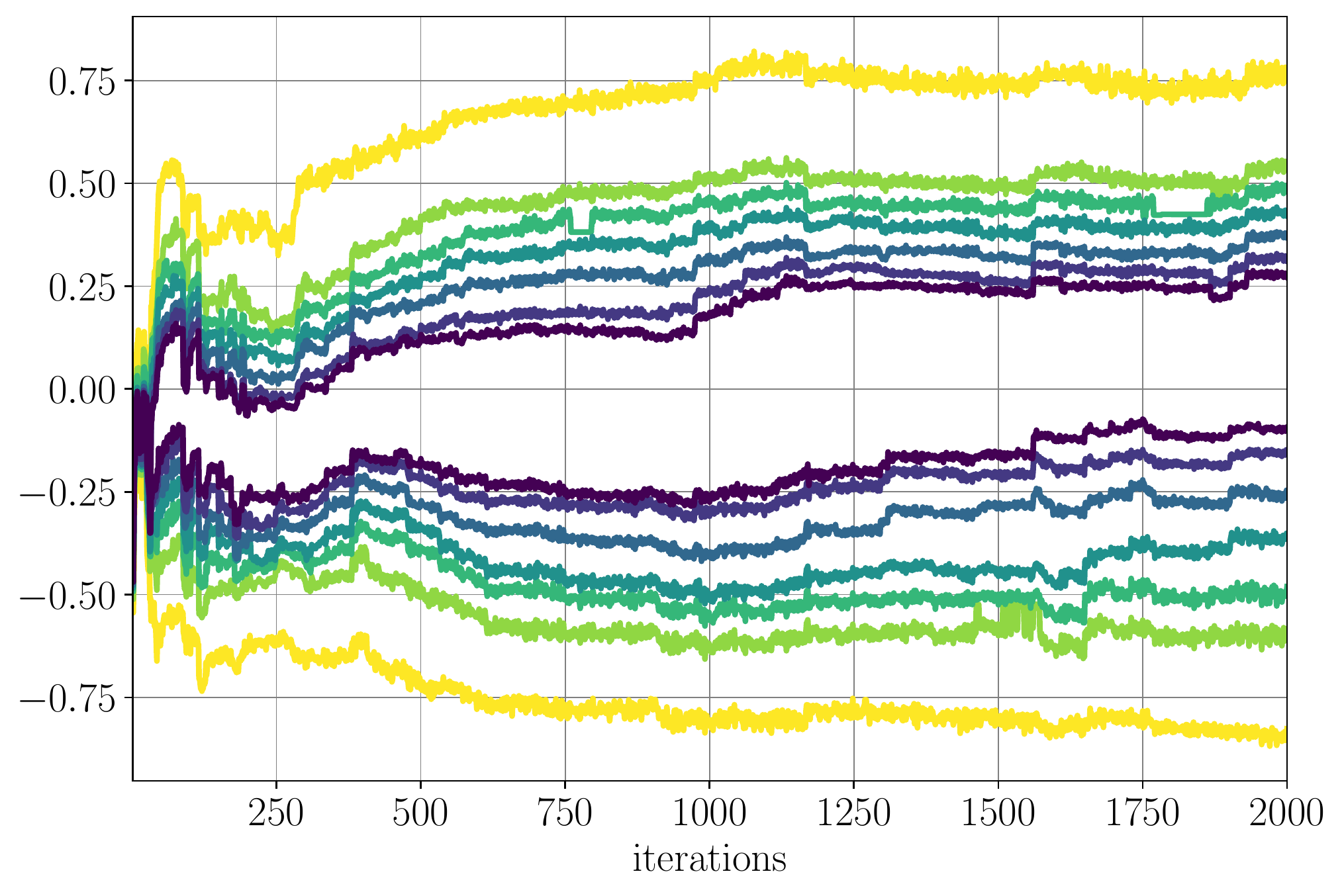}%
  \hfill
  \includegraphics[width=\fracwidth]{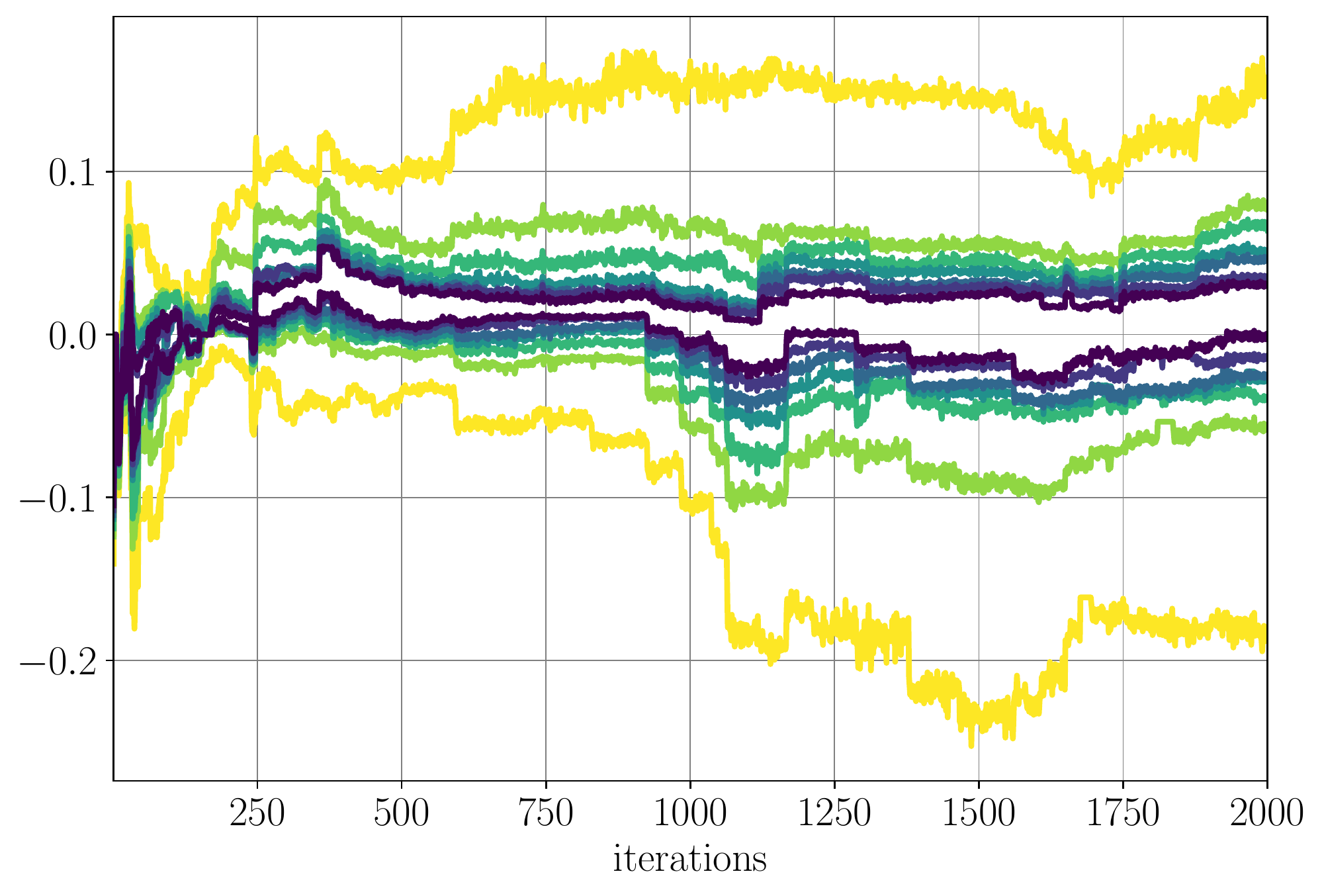}%
  \caption{We show results of Gunsilius's algorithm for 3 different settings of $y^{\star} \in \{0.25, 0.5, 0.75\}$.}
  \label{fig:gunsilius}
\end{figure}

\section{The Shape of \texorpdfstring{$p_\eta(\theta \given x, z)$}{p(theta | x, z)} and Conditional Effects}
\label{app:p_eta}

It is not difficult to show that our parameterization of $p_\eta(\theta \given x, z)$ enforces $\theta \indep Z$ while allowing for $\theta \dep Z \given X$, as suggested by Figure~\ref{fig:setup}(c).
It follows directly by factoring a conditional density in terms of a copula density $c(\cdot)$ and the required univariate marginals.
That is, for some $(V_1, V_2, V_3)$ for which we want to define a conditional pdf $p(v_2 \given v_1, v_3)$, we have
\begin{align*}
  p(v_1, v_2 \given v_3) &:= c(F(v_1 \given v_3), F(v_2 \given v_3))\, p(v_1 \given v_3)p(v_2 \given v_3)\quad \Rightarrow\\
  p(v_2 \given v_1, v_3) &\phantom{:}= c(F(v_1 \given v_3), F(v_2 \given v_3))\, p(v_2 \given v_3).
\end{align*}
Since $\int p(v_1, v_2 \given v_3)\, dv_1 = p(v_2 \given v_3)$, a necessary and sufficient condition for $V_2 \indep V_3$ is choosing a model marginal such that $p(v_2 \given v_3) = p(v_2)$.
If $c(F(v_1 \given v_3), F(v_2))$ cannot be factored in terms of some product $h_1(v_1, v_3)h_2(v_1, v_2)$, which is typically the case, then $V_2 \dep V_3 \given V_1$.

The main apparent limitation of our $p_\eta(\theta_k)$ (and the related copula) is its reliance on a parametric form.
There is a complex relationship between the shape of the response function space and the distribution implied on that space by the unknown model $\mathcal{M}$.
For $Y = f(X, U)$, it is always possible to assume without loss of generality that $U$ is a set of variables which are marginally standard Gaussians: just let the transformation $U'_i := \Phi^{-1}(F_i(U_i)))$ be absorbed into $f(\cdot)$, where $F_i(\cdot)$ is the marginal CDF of $U_i$ and $\Phi(\cdot)$ is the CDF of a standard Gaussian.
Moreover, assuming that any dependence among elements of $U$ can be explained by direct causation among them or by other latent parents, we can also assume all members of $U$ are independent.

However, we do not want to assume a one-to-one correspondence between elements of $\theta$ and elements of $U$: that is the whole point of using response functions.
Even independent standard Gaussian $U$s would not translate to marginally Gaussian $\theta$.
As an example, suppose $Y = U_1^2X + \lambda U_2$.
All response functions can be written in the form $f_\theta(x) := \theta_1x + \theta_2$, where $\theta_1 = U_1^2$ and $\theta_2 = \lambda U_2$.
Hence, $\theta_1$ follows a chi-squared distribution and $\theta_2$ a zero-mean, but not standard, Gaussian.
If $Y = U_1X^2 + \lambda U_1U_2$, then on top of that $\theta_1$ and $\theta_2$ are not independent.

The solution is conceptually not complicated: just let $p_\eta(\cdot)$ be as flexible as desired.
For instance, \emph{let the copula be a finite or Dirichlet process mixture of Gaussian copulas, also defining flexible models for the marginals}.
The IV conditional independence structure among $Z, X, \theta$ is still preserved.
The practical issue of course is the optimization.
The algorithm of \cite{gunsilius2019bounds} itself tries to approach the problem by learning the reweighting of a Monte Carlo approximation to a fixed base measure.
That alone is already very computationally demanding and has convergence problems. 

We set a parametric form for $p_\eta(\cdot)$ for reasons beyond a compromise between flexibility and computational tractability.
\emph{Adopting a nonparametric model for the causal model, such as a Dirichlet process, seems pointless because:}
(a) we do not perform statistical inference directly in the causal model, but only via  black-box estimators of (features of) $p(x, y \given z)$, which can be nonparametric;
(b) if we were to follow the route of performing statistical inference by directly fitting the causal model, the corresponding estimator would have a finite representation with dimensionality given by the data.
A practical resource, sample size, limits the representational size of the estimator.
The role of nonparametrics is to provide a type of adaptive regularization, and to provide theory about limits of parametric estimators as done by \cite{gunsilius2019bounds}.
The latter has clear value in itself but it does not demand nonparametric models to be actually implemented, while the former is out of our scope: in our case, no regularization is needed for the causal model as we do not fit data based on it.
Instead, our practical resource is the computational budget: if we want to not use domain knowledge to perform the causal analysis, we simply choose the size of $\eta$ based directly on the main bottleneck, the amount of computation available.
Hence, by the time-data bounded nature of computational and statistical inference, we lose nothing by adopting a finite representation for both $\eta$ and $\theta$. 

The practitioner should be invited to sample from the implied function space to visualize whether the distribution of sample paths has a desired level of variability.
Getting the ``exact'' shape of the true distribution is however nowhere as important as just having enough variability to avoid overconfident bounds.
How to achieve ``enough variability'' without aiming at a completely flexible distribution of $\theta$ may be a compromise between computational costs and domain-dependent judgment.
In particular, given the choice of the $\{\phi_l\}$ family by which we link the causal model to observation, we may opt for the \emph{maximum entropy distribution} that is given by the corresponding moments, the Gaussian in case of first and second moments of $p_\eta(\theta \given x, z)$---although this still leaves open how the mean and covariance matrix of $\theta$ will change with $x$ and $z$.

In any case, the finite mixture of Gaussians approach can still be implemented with the reparameterization trick.
The relation to Gunsilius algorithm is that our ``base measure'' is smoothly adaptive, leading to possibly more stable behavior in practice.
The price to be paid is that each iteration in our method would be substantially more expensive than the efficient mixture component weighting optimization done at each iteration of Gunsilius' method, \emph{if} we were to optimize the mixture component parameters to completion while fixing the samples.
However, we  do joint partial optimization by gradient-informed small steps, taken at each sampling stage.
This is one of the main distinctive features of our class of algorithms compared to the resample/optimize alternating procedure of \cite{gunsilius2019bounds}.

To summarize, \emph{the Gaussian case, discussed in the main text, should be seen as a useful illustration, not as a one-size-fits-all solution}.
Any copula for which the reparameterization trick can be used can be automatically plugged into any instance of our class of algorithms.

Another important aspect brought by a parameterization of $p_\eta(\cdot)$ is in case we have pre-treatment covariates $W$ to either reduce confounding, remove (direct) dependence between $Z$ and $U$ or $Z$ and $Y$, or just to answer questions related to conditional expected outcomes, e.g., $\E[Y \given do(x), w]$ and conditional average causal effects (CATE), $\E[Y \given do(x), w] - \E[Y \given do(x'), w]$.
Although a response function can straightforwardly depend on a vector of treatment variables, this makes less sense if variables $W$ are not direct causes of $Y$.
And even if elements of $W$ are direct causes, we may want to treat them analogously to $U$: playing a role in the response function only via the distribution of $\theta$, instead of being explicitly in the scope of such functions.

\emph{Modeling CATE can then be done in a completely straightforward way.}
Nothing in the algorithm changes if we use a probabilistic model for $p(x, y\given z, w)$ to provide the observable counterpart of the causal model.
Each configuration $w$ defines a separate optimization problem.
The corresponding factor $p(\theta \given x, z, w)$ can be set independently for each instance of $w$, regardless of its dimensionality.

However, a practitioner may be interested on providing information about how $p(\theta \given x, z, w)$ varies smoothly across values of $w$ in order to impose further constraints on the response functions across multiple $w$ realizations.
We suggest that a way of incorporating covariates $W$ is by a multilevel approach: define $p_{\eta(w)}(\theta \given x, z, w)$, where each element of $\eta$ may itself be a function of $W$, e.g., $\mu_1 = \beta_1^\mathsf{T}W$ for some parameter vector $\beta_1$.
Here, $p(x \given z, w)$ and $p(y \given z, w)$ (or $p(y \given x, z, w)$) are the marginals to be matched.
We will discuss in future work ways of making $p_\eta(\cdot)$ more flexible in general, including the use of covariates.

\section{Discrete Outcomes and Discrete Features}
\label{app:discrete_outcomes}

If $Y$ is discrete, $f_\theta(x)$ will be discontinuous.
Theoretically this will not pose a problem as long as the number of discontinuities is finite \citep{gunsilius2019bounds}.
The main practical issue is optimization, as eq.~(\ref{eq:constraints_approx2}) will now not lead itself to gradient-based methods.
The most immediate approximation is to use differentiable surrogates of $f_\theta(x)$ that relax the constraints.
In the most basic formulation, we have the inequalities
\begin{equation*}
 tol_{-} \leq \E[\phi_l(Y) \given z^{(m)}] - \int \phi_l(f_{\theta}(x)) \, p_\eta(x, \theta \given z^{(m)})\, dx~d\theta \leq tol_{+},
\end{equation*}
for some tolerance factors $tol_+, tol_-$.
Given upper and lower bounds $\phi_l^+(f_{\theta}(x))$,  $\phi_l^-(f_{\theta}(x))$ on  $\phi_l(f_{\theta}(x))$, the relaxed constraints
\begin{align*}
 tol_{-} \leq {} & \E[\phi_l(Y) \given z^{(m)}] - \int \phi_l^-(f_{\theta}(x)) \, p_\eta(x, \theta \given z^{(m)})\, dx~d\theta \\
 & \E[\phi_l(Y) \given z^{(m)}] - \int \phi_l^+(f_{\theta}(x)) \, p_\eta(x, \theta \given z^{(m)})\, dx~d\theta \leq tol_{+},
\end{align*}
will still result in valid, but looser bounds (again, up to local optima and Monte Carlo error).
If $f_\theta(x)$ is non-negative (for instance, if its codomain is $\{0, 1\}$) and $\phi_l(\cdot)$ is monotonic for non-negative inputs (such as $\phi_l(x) = x$ and $\phi_l(x) = x^2$), it is enough to plug in bounds for $f_\theta(x)$ itself.
We will elaborate on that in future work.
In this context, we can also formulate an alternative approach to matching $p(y \given z)$.

\xhdr{Alternative Approach to Matching \texorpdfstring{$p(y\given z)$}{p(y | z)}}
Here we describe an alternative approximation of eq.~\eqref{eq:constraints_exact} that hinges on smoothly approximating the indicator function to render the integral well behaved.
First, instead of evaluating $\Pr(Y < y \given Z = z^{(m)})$ for all $y \in \cY$, we take a similar approach for discretizing $Y \given z^{(i)}$ as we took for $z^{(m)}$.
For a given $z^{(m)}$, instead of all half-spaces $Y < y$, we only consider the sets
\begin{equation*}
  A^{(m, l)} := (-\infty, y^{(m, l)}] \quad \text{with} \quad y^{(m, l)} := F_{Y \given z^{(m)}}^{-1}\Bigl(\frac{l-1}{L-1}\Bigr)
\end{equation*}
for $l \in [L]$ with some fixed $L \in \bN$.
This results in constraints for the $L$-quantiles of the conditional distributions of $Y$
\begin{equation*}
  \frac{l-1}{L-1} = \int \B{1}\left(f_{\theta}(x) \le y^{(m, l)}\right) \, p_\eta(x, \theta \given z^{(m)})\,dx~d\theta.
\end{equation*}
for all $m \in [M]$ and $l \in [L]$.
In practice, we would evaluate the integral on the right hand side with a Monte Carlo estimate, sampling from $p_{\eta}(x, \theta \given z^{(m)})$ and then differentiate with respect to $\eta$ for gradient-based optimization.
Therefore, the non-differentiable (even non-continuous) indicator function poses an issue for the optimization.
We can circumvent this problem by approximating the indicator with a smoothly differentiable function, for example
\begin{equation*}
  \B{1}(t \le t^*) \approx \sigma_{\rho}(t - t^*) \quad \text{for} \quad \sigma_{\rho}(t) := \frac{1}{1 + e^{-\rho\, t}} \quad {\color{gray}\text{or} \quad \sigma_{\rho}(t) := \frac{1}{1 + \exp\Bigl(-\rho\, \bigl(t + \frac{1}{\sqrt{\rho}}\bigr)\Bigr)}}
\end{equation*}
for $\rho > 0$.
As $\rho \to \infty$, $\sigma_{\rho}(t) \to \B{1}(t \le 0)$ pointwise on $\bR \setminus \{0\}$, i.e., we can slowly increase $\rho$ throughout the optimization to gradually approximate the constraints.

Hence an alternative approach to implement the constraint for matching $p(y \given z)$ is
\begin{equation*}
  \frac{l-1}{L-1} = \int \sigma_{\rho}\bigl(f_{\theta}(x) - y^{(m, l)}\bigr) \, p_\eta(x, \theta \given z^{(m)})\,dx~d\theta
\end{equation*}
for all $m \in [M]$ and $l \in [L]$, where we increase $\rho > 0$ after each optimization round.

In practice, we this approach gave less robust results than the approach described in the main text, partly due to the additional hyperparameter schedule needed for $\rho$.
Therefore, we only report results for the approach using dictionary functions $\phi_l$ described in the main text.

\section{Algorithm}
\label{app:algo}

\begin{algorithm}[t!]\label{alg:optimization}
\caption{Bounding the IV interventional effect at treatment level $x^{\star}$.}
\begin{algorithmic}[1]
  \Require
  dataset $\cD = \{(z_i, x_i, y_i)\}_{i=1}^N$;
  number of $z$ grid points $M$;
  constraint functions $\{\phi_l\}_{l=1}^L$;
  response function family $\{f_{\theta}\}_{\theta \in \Theta}$;
  batchsize $B$;
  initial temperature $\tau^{(0)} > 0$;
  temperature increase factor $\alpha > 1$;
  tolerances $\abstol, \reltol$;
  initial Lagrange multipliers $\lambda$;
  initial parameters $\eta^{(0)}$;
  \State $z^{(m)} := \hat{F}_Z^{-1}(\frac{m}{M + 1})$ for $m \in [M]$\Comment{$\hat{F}_Z$: CDF of $\{z_i\}_{i=1}^N$.}
  \State $\bin(i) := \max\{\argmin_{m \in [M]} |z_i - z^{(m)}|\}$ for $i \in [N]$\Comment{split data points into ``z-bins''}
  \State $\lhs_{m, l} := \frac{1}{|\bin^{-1}(m)|} \sum_{i \in \bin^{-1}(m)} \phi_l(y_i)$ for $m \in [M], l \in [L]$\Comment{pre-compute $\lhs$}
  \State smoothen $\lhs_{m,l}$ across $m$ for each $l$ with spline regression\Comment{see Appendix~\ref{app:smoothing}}
  \State $b := \max\{\abstol, \reltol \, \lhs\}$ (element-wise)\Comment{set constraint tolerances}
  \State $\hat{x}_j^{(m)} := \hat{F}_{X\given z^{(m)}}^{-1}\Bigl(\frac{j-1}{B-1}\Bigr)$ for all $j \in [B], m \in [M]$\Comment{$\hat{F}_{X\given z^{(m)}}$: CDF of $\{x_i\}_{i \in \bin^{-1}(m)}$}
  \For{$t = 1 \ldots T$ (or until convergence)}\Comment{optimization rounds}
  \State $\eta^{(t)} := \Call{OptimizeSubproblem}{\eta^{(t - 1)}, \lambda^{(t - 1)}, \tau^{(t - 1)}}$\Comment{min.\ Lagrangian at fixed $\lambda, \tau$}
  \State $\lambda_l^{(t)} \gets \max\left( 0, \lambda_l^{(t-1)} - \tau^{(t-1)} c_l(\eta^{(t)})\right)$\Comment{update Lagrangian multipliers}
  \State $\tau^{(t)} \gets \alpha\, \tau^{(t-1)}$\Comment{increase temperature parameter}
  \EndFor
  \State \Return $o_{x^{\star}}(\eta^{(T)})$
  \Statex
  \Function{OptimizeSubproblem}{$\eta, \lambda, \tau$}
    \State \Comment{In here we use SGD with auto-differentiation to minimize $\cL$. Hence we only describe how to evaluate $\cL$ in a differentiable fashion:}
    \State $o_{x^{\star}}(\eta) := \frac{1}{B} \sum_{j = 1}^B f_{\theta^{(j)}}(x^{\star})$ with $\theta^{(j)} \sim p_{\eta}(\theta)$\Comment{c.f.~Algorithm~\ref{alg:sampling} for sampling}
    \State $\rhs_{m,l}(\eta) := \frac{1}{B} \sum_{j = 1}^B \phi_l\bigl(f_{\theta^{(j)}}(\hat{x}^{(m)}_j)\bigr)$\Comment{c.f.~Algorithm~\ref{alg:sampling} for sampling}
    \State $c(\eta) := b - |\lhs - \rhs(\eta)|$\Comment{compute constraint terms}
    \State $\cL(\eta) := \B{\pm} o_{x^{\star}}(\eta) + \sum_{l=1}^{M\cdot L} \xi(c_l(\eta), \lambda_l, \tau)$\Comment{Lagrangian ($\pm$ for lower/upper bound)}
    \State \Return $\argmin_{\eta} \cL(\eta)$\Comment{optimize with SGD}
  \EndFunction
\end{algorithmic}
\end{algorithm}

\subsection{Additional Details of the Optimization}
\label{app:smoothing}

\xhdr{Smoothen $\lhs$}
Since $\lhs_{m, l}$ are estimated via empirical averages of $\phi_l(y_i)$ for datapoints in a given bin $i \in \bin^{-1}(m)$, ``neighboring'' constraints $\lhs_{m,l}$ and $\lhs_{m+1,l}$ may have substantially different values.
Since our model is smooth, it can be hard to match such non-continuities with $\rhs_{m,l}(\eta)$.
Intuitively, we expect such jumps to be artifacts of finite sample effects and not important properties of the true data distribution.
Hence we apply a spline regression to the values $\{\lhs_{m,l}\}_{m=1}^M$ for each $l \in [L]$ to smoothen out larger jumps between neighboring values.
In practice, we use a cubic univariate spline for each $l$ with a smoothing factor of $0.2$.

\subsection{Augmented Lagrangian Optimization Strategy}

The Augmented Lagrangian method \citep{hestenes1969multiplier} is a general method for constrained optimization, originally proposed just for dealing with equality constraints.
The benefit of this over penalty methods is that we do not need to take the penalty parameters $\tau$ to $\infty$ in order to solve the original constrained optimization problem, which can cause ill-conditioning \citep{nocedal2006numerical}.
However, our problem only contains inequality constraints.
Thus, we consider a refinement proposed by \cite{nocedal2006numerical} to purely handle inequality constraints using Augmented Lagrangian methods.
Specifically, we can write the inequality constrained optimization problem equivalently as an unconstrained optimization problem with Lagrange multipliers $\lambda$:
\begin{equation*}
  \min_{\eta} \max_{\lambda \geq 0} \Big\{ o_{x^{\star}}(\eta) + \lambda^\top (c(\eta) - b) \Big\}.
\end{equation*}
To see that it is equivalent, note that the $\max$ returns $o(\eta)$ when $\eta$ satisfies the constraints (as the maximum is obtained at $\lambda = 0$), and $\infty$ otherwise (as the maximum is at $\lambda = \infty$).
However, this is not easy to optimize as the $\lambda$ jumps from $0$ to $\infty$ when passing through the constraint boundary.
To fix this, we add a term that penalizes $\lambda$ making larger changes from its previous value.
Specifically,
\begin{equation*}
  \min_{\eta} \max_{\lambda \geq 0} \Big\{ o(\eta) + \lambda^\top (c(\eta) - b) - \frac{1}{2 \tau} \| \lambda - \lambda' \|^2 \Big\},
\end{equation*}
where $\lambda'$ are the Lagrange multipliers from the previous iteration and $\tau$ is a penalty term that is iteratively increased.
Note that the $\max$ optimization can be solved in closed form for each Lagrange multiplier $\lambda_l$
\begin{equation*}
  \lambda_l = \max\left\{0, \lambda'_l + \tau c_l(\eta)\right\},
\end{equation*}
where $c_l(\eta)$ is shorthand for the $l$-th inequality constraint.
Plugging these values into the optimization problem, we arrive at
\begin{equation*}
  \min_{\eta} \cL(\eta, \lambda, \tau) := o_{x^{\star}}(\eta) + \sum_{l=1}^{M \cdot L} \xi(c_l(\eta), \lambda_l, \tau)
\end{equation*}
with
\begin{equation*}
  \xi(c_l(\eta), \lambda_l, \tau) :=
  \begin{cases}
    - \lambda_l c_l(\eta) + \frac{\tau c_l(\eta)^2}{2}
      & \text{if } \tau c_l(\eta) \le \lambda_l,  \\
    - \frac{\lambda_l^2}{2 \tau}
      & \text{otherwise},
  \end{cases}
\end{equation*}
where $\tau$ is increases throughout the optimization procedure.
Given an approximate solution $\eta$ of this subproblem, we then update $\lambda$ according to
\begin{equation*}
  \lambda_l \gets \max\{0, \lambda_l - \tau c_l(\eta)\}
\end{equation*}
for all $l \in [M \cdot L]$ and set $\tau \gets \alpha\, \tau$ for a fixed $\alpha > 1$.
For the full optimization, we attach temporal upper indices, i.e., at time step $t$, we have the current approximate solution $\eta^{(t)}$, the Lagrange multipliers $\lambda^{(t)}_l$ and the temperature parameter $\tau^{(t)}$.
See Algorithm~\ref{alg:optimization} for a description of the optimization scheme.
While the number of optimization parameters grows quickly with the dimensionality of $\theta$, which may render the optimization challenging, in our experiments we did not encounter any issues with up to 54 optimization parameters and 40 constraints.

\begin{algorithm}[t!]
\caption{Sampling parameter values $\theta$ from $p_{\eta}(\theta, X \given z^{(m)})$.}\label{alg:sampling}
\begin{algorithmic}[1]
  \State Sample each component of $w \in \bR^{K \times B}$ i.i.d.\ from a standard Gaussian.
  \State Prepend the vector $(0, 1/B, \ldots, 1)$ as the first row of $w$, resulting in $w \in \bR^{(K+1) \times N}$.
  \State Allow for dependencies between components by multiplying with the Cholesky factor $w \gets L\, w$.
  \State Normalize all values by applying the standard Gaussian CDF component wise, $w \gets \varphi_{0, 1}(w)$.
  \State Fix the marginals of $\theta_k^{(j)}$ by applying the inverse CDF of a $(\mu_k, \sigma_k^2)$-Gaussian: $\theta^{(j)} \gets \varphi_{\mu_k, \sigma_k^2}^{-1}(w_{j+1})$ for $j \in [K]$.
  Here, $w_{j+1}$ denotes the $j+1$-st row of $w$.
  \State {\color{gray}Sampling $X$ via $\hat{F}_{X}^{-1}(w_1)$ by design simply gives the pre-computed $\hat{x}$.}
\end{algorithmic}
\end{algorithm}

\subsection{Sampling from the Copula}

A crucial step for our algorithms was baking the assumptions about $p(X \given Z)$ as well as $Z \indep U$ directly into our model from which we sample for Monte Carlo estimates.
Algorithm~\ref{alg:sampling} describes in detail how we can obtain these samples from the copula defined in eq.~\eqref{eq:copula} in a differentiable fashion with respect to $\eta$.

\subsection{Parameter Initialization}
\label{app:initialization}

We initialize the optimization parameters $L$ with ones on the diagonal, zeros in the upper triangle, and sample the lower triangle from $\cN(0, 0.05)$.
The initialization for $\mu_k$ and $\ln(\sigma_k^2)$ depends on the chosen response function family.
Our guiding principle is to ensure that the initial distribution covers a large set of possible response functions, tending towards larger $\sigma_k$.

\section{Response Functions}
\label{app:response}

One key advantage of our approach is that it allows us to flexibly trade off assumptions on the response function family with more informative bounds.
Due to our simple, yet expressive choice of linear combinations of a set of basis functions, there are many natural and easy to implement options for the response functions.
In particular, we consider the following options:
\begin{enumerate}[leftmargin=*]
  \item \emph{Polynomials:} $\psi_k(x) = x^{k - 1}$ for $k \in [K]$.
  In this work, we specifically focus on linear ($K = 2$), quadratic ($K = 3$), and cubic ($K = 4$) polynomial functions.
  \item \emph{Neural basis functions (MLP):} We fit a multi-layer perceptron with $K$ neurons in the last hidden layer to the observed data $\{(x_i, y_i)\}_{i \in N}$ and take $\psi_k$(x) to be the activation of the $k$-th neuron in the last hidden layer.
  Note that the network output itself is a linear combination of these last hidden layer activations.
  Hence, the underlying assumption for this approach to work well is that the true causal effects can also be approximated well by a linear combination of the learned last hidden layer activations, i.e., the true effect is in this sense ``similar'' to the estimated observed conditional $\hat{p}(y \given x)$.
  In practice, we train a 2-hidden layer MLP with 64 neurons in each layer, rectified linear units as activation functions and an mean-squared-error loss for 100 epochs and a batchsize of 256 using Adam with a learning rate of $0.001$.
  \item \emph{Gaussian process basis functions (GP):} We fit a Gaussian process with a sum-kernel of a polynomial kernel of degree $3$, an RBF kernel, and a white noise kernel to $K$ different sub-samples $\{(x_i, y_i)\}_{i \in N'}$ with $N' \le N$.
  We then sample a single function from each Gaussian process as the basis functions $\psi_k$ for $k \in [K]$.
  We train multiple Gaussian processes on smaller subsets of the data to ensure sufficient variance in the learned functional relation.
  Similarly to the neural net basis functions, the assumption is that the causal effect can be approximated by a linear combination of these varying samples.
  In our experiments, we fit the Gaussian processes with scikit-learn's \texttt{GaussianProcessRegressor} \citep{scikit-learn} using $N' = 200$ and a white kernel variance of $0.4$.
\end{enumerate}

\section{Why Discretization is not a Good Idea}
\label{app:discretization}
 
The framework of \cite{balke1994counterfactual} is powerful and simple, and hence it raises the prospect that discretizing treatment $X$ can provide a good approximation to the original problem where $X$ is continuous.
However, there are several reasons why this is not a good idea:
\begin{itemize}
\item \emph{It destroys the key assumption of instrumental variable modeling}.
  Besides the lack of confounding between instrument $Z$ and outcome $Y$, the key assumption in an IV model is the conditional independence $Y \indep Z \given \{X, U\}$ (``exclusion restriction'').
  This assumption will in general fail to hold if we destroy information, i.e., if we condition on $X \in \cA$, for some set $\cA$, instead of the realization of $X$;
\item \emph{It makes causal estimands ill-defined.}
  There are several ways in which an intervention can be ambiguous.
  This happens when defining the manipulation of a construct (``race'') or of summary measurements in general (``obesity'').
  One particular instance of the latter is when we speak of $do(x^\star)$, meaning the setting of a discretization $X^\star$ of $X$ to a particular level $x^\star$ \citep{vanderweele:13}.
  If $X^\star = x^\star$ corresponds to the event $X \in [a, b]$, then this at least needs the assumption that $\E[Y \given do(x)]$ is approximately constant for $x \in [a, b]$ for the intervention to be meaningful.
  This is pointless if the goal is to avoid making assumptions about the shape of the response function;
\item \emph{Its cost is super-exponential.}
  Suppose we still want to proceed with the idea of discretization, in the sense that we are willing to assume that we are using a fine enough grid of intervals for the treatment so that the previous two points are not particularly prominent.
  It may be argued that using
  \cite{balke1994counterfactual} with this approximation is attractive on the grounds it is a convex, deterministic approach and hence a more computationally attractive alternative to tackling the continuous problem.
  In fact, the opposite may hold. Assume we discretize $X$ and $Y$ to $|\cX|$ and $|\cY|$ levels respectively, and $Z$ assumes $|\cZ|$ levels (perhaps also by discretization). Then the cost of using the full information of the distribution is approximately $\cO(|\cX|^{|\cZ|} |\cY|^{|\cX|})$.
  It is true that, just like in our approach, this can be much simplified if we rely only on a subset of constraints.
  In particular, if we use only the first moments in the constraints and the expected outcome is the objective function, we can simplify the discrete formulation by targeting our parameterization to depend only on the expected outcomes directly.
  This makes the problem exponential only on $|\cZ|$, see for instance the parameterization of \cite{zhang:20}.
  Being ``only'' exponential may still require Monte Carlo approximations in general.
  But this can still be super-exponential if $|\cZ|$ grows with $|\cX|$, which will be necessary if the instrument is strong:
  for an extreme example, if $Z$ and $X$ lie close to a line with high probability and we choose only two levels of $Z$ against many levels of $X$, then most combinations of pre-determined $(z, x)$ pairs will lie on regions of essentially zero density in the $p(x, z)$ distribution;
\item \emph{It is vacuous in the limit}.
  Even if we can use an arbitrarily fine discretization and assume that the piecewise nature of the approximation is close enough to the true response functions of $Y$, we know that as $|\cX| \rightarrow \infty$ the number of discontinuities in the response function also goes to infinity.
  As described by \cite{gunsilius2018testability}, we will not learn anything non-trivial about the causal estimand of interest.
\end{itemize}
 
We reiterate the points above in more direct way:
\emph{being unable to express constraints on the response function is not an asset, it's a liability.}
Discretization allows us to easily use a single family of functional constraints: piecewise constant functions.
In this framework, it is cumbersome to represent other constraints such as smoothness constraints, and \emph{the degree of violation of the exclusion restriction assumption remains unknown}.
There is no reason to believe this discrete representation is a good family in any computationally bounded sense, as an efficient choice of discretization points can only be made if we know something about the function.
And if we do, then it makes far more sense to use more representationally efficient ways of partitioning the space of $X$, such as regression splines with a fixed number of knots.
This involves no discretization of treatment, while avoiding the issues of violation of the exclusion restriction assumption and ambiguity of intervention.

\section{Modeling \texorpdfstring{$p(y \given x, z)$}{p(y | x, z)} and Monte Carlo Alternatives}
\label{app:p_yxz}

Alternatively to the setup described in the main text, we can match not only the marginal $p(y \given z)$, but the theoretically more informative $p(y \given x, z)$.
This problem is actually conceptually simpler, although it will require joint measurements over the three types of variables. 

The main modification is as follows.
Instead of
\begin{equation*}
  \E[\phi_l(Y) \given Z = z^{(m)}] = \int \phi_l(y_{\theta}(x)) \, p_\eta(x, \theta \given Z = z^{(m)})\, dx~d\theta,
\end{equation*}
we build constraints based on
\begin{equation*}
  \E[\phi_l(Y) \given X = x^{(m)}, Z = z^{(m)}] = \int \phi_l(y_{\theta}(x^{(m)})) \, p_\eta(\theta \given X = x^{(m)}, Z = z^{(m)})\, d\theta,
\end{equation*}
where now we need to define a grid over the joint space of $X$ and $Z$.
This can be done in several ways, including the joint product of equally-spaced quantiles of the respective marginal distributions, perhaps discarding combinations for which $p(x^{(m)}, z^{(m)})$ are below some threshold.
Moreover, the factor $p_\eta(\theta \given X = x^{(m)}, Z = z^{(m)})$ was explicitly parameterized  in our original setup, and can be used as is.

Notice the advantages and disadvantages of the two approaches.
Modeling the full conditional $p(y \given x, z)$ uses the full information of the problem (as it is equivalent to $p(x, y \given z)$, where $p(x \given z)$ is tackled directly), which in principle is more informative but requires functionals of the joint $p(x, y \given z)$ instead of the marginals $p(x \given z)$ and $p(y \given z)$.
We can also see that we are trading-off adding more constraints but removing the need to integrate $X$ in each constraint.
More interestingly, this full conditional approach does not require any kind of density estimation: the need for $p(x \given z)$ disappears, and all we need on the left-hand sides are estimates of expectations.

More generally, it is clear that there are practical cases where $\E[\phi_l (y_{\theta}(X)) \given x, z]$, with $\theta$ being the random variable to be marginalized, has an analytical solution as a function of $\eta$.
For instance, this will be the case when $\phi_l$ stands for linear and quadratic functions of $Y$ (itself a linear function of $\theta$ for a fixed $x$), and $p_{\eta}(\theta \given x, z)$ is a (mixture of) Gaussian(s), which is true for our experiments.
However, we demonstrate the suitability of Monte Carlo formulation in order to provide and evaluate a class of algorithms for black-box (differentiable) features, where such expectations cannot be computed analytically in general. 

\section{Fitting Latent Variable Models}
\label{app:latent}

When fitting the latent variable models, we use multi-layer perceptrons with inputs $z, x, y$ for the means and variances of the latent dimensions $U$, where we use lower indices $U_i$ for the different components.
For this encoder, we use 32 neurons in the hidden layer and rectified linear units as the activation function.
There are two decoders.
The first one is trained to reconstruct $\E[X \given X, U]$, i.e., receives the original $Z$ in addition to the latent vector $U$ as input.
It is also parameterized by an MLP with 32 neurons in the hidden layer and ReLu activations.
The second decoder reconstructs $\E[Y \given X, U]$ and is either an MLP of the same architecture (when comparing to MLP response functions), linear in $X$, i.e., $\alpha X + \beta + \sum_{i =1}^{\mathrm{n\_latent}} (\gamma_i X U_i + \delta_i U_i)$ (when comparing to linear response functions), or quadratic in $X$, i.e., $\alpha X^2 + \beta X + \gamma + \sum_{i =1}^{\mathrm{n\_latent}} (\delta_i X^2 U_i + \epsilon_i X U_i + \zeta_i U_i)$ (when comparing to quadratic response functions).
Thereby, we ensure that the form of  matches our assumptions on the function form of the response family.
We then optimize the evidence lower bound following standard techniques of variational autoencoders \citep{kingma2013auto} with $L_2$ reconstruction loss for $X$ and $Y$.
We fit multiple models with different random initializations and compute the implied causal effect of $X$ on $Y$ for each one, which is obtained from the decoder $\E[Y \given u, x]$ by averaging over 1000 samples of the latent variable $U$ for a fixed grid of $x$-values.

\section{Additional Experimental Results}
\label{app:results}

\def\fracwidth{0.33}
\def\spacer{\hspace{2.6cm}}
\begin{figure}
  \centering
  cubic response\spacer GP response \spacer MLP response\\
  \includegraphics[width=\fracwidth\textwidth]{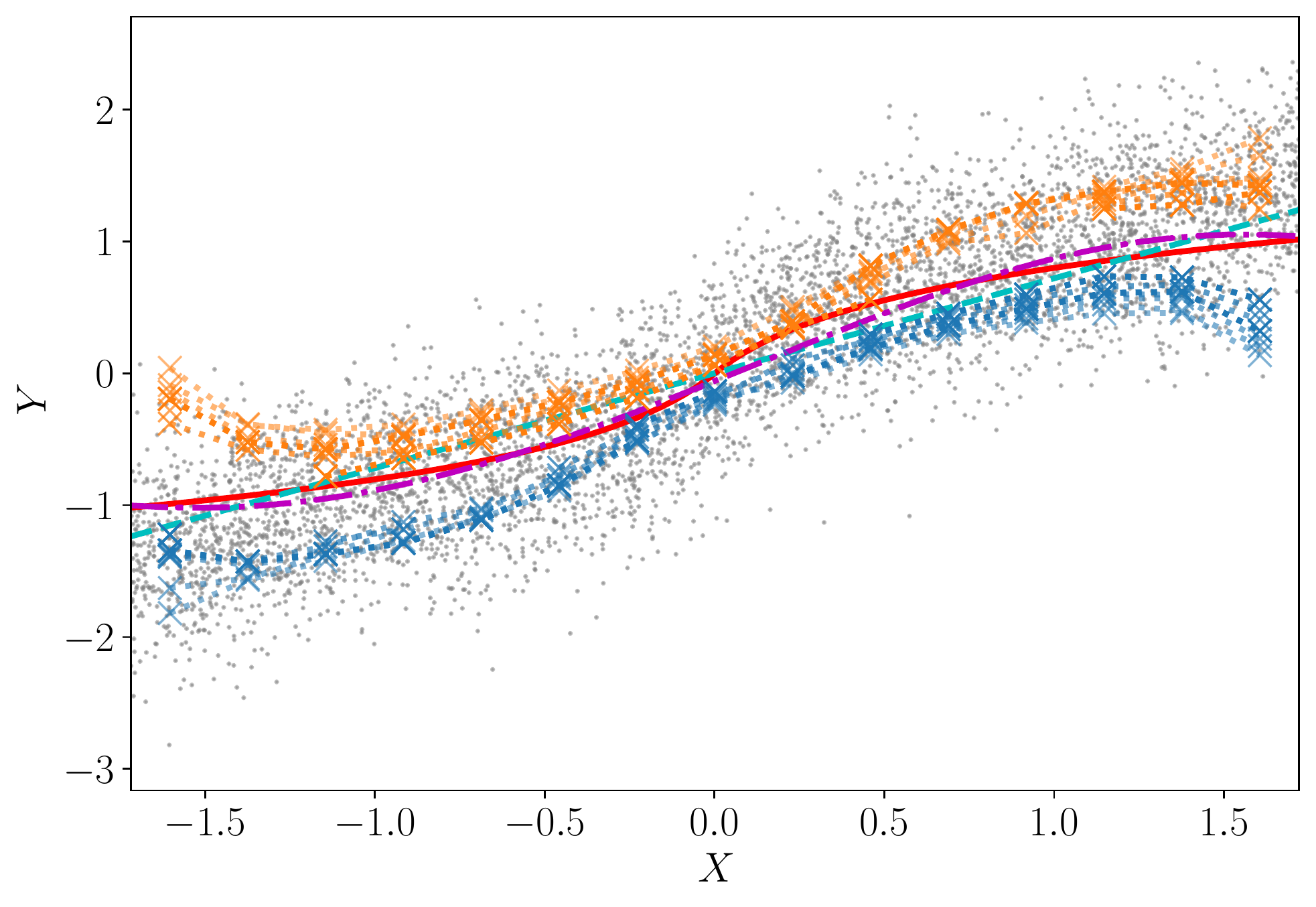}%
  \hfill
  \includegraphics[width=\fracwidth\textwidth]{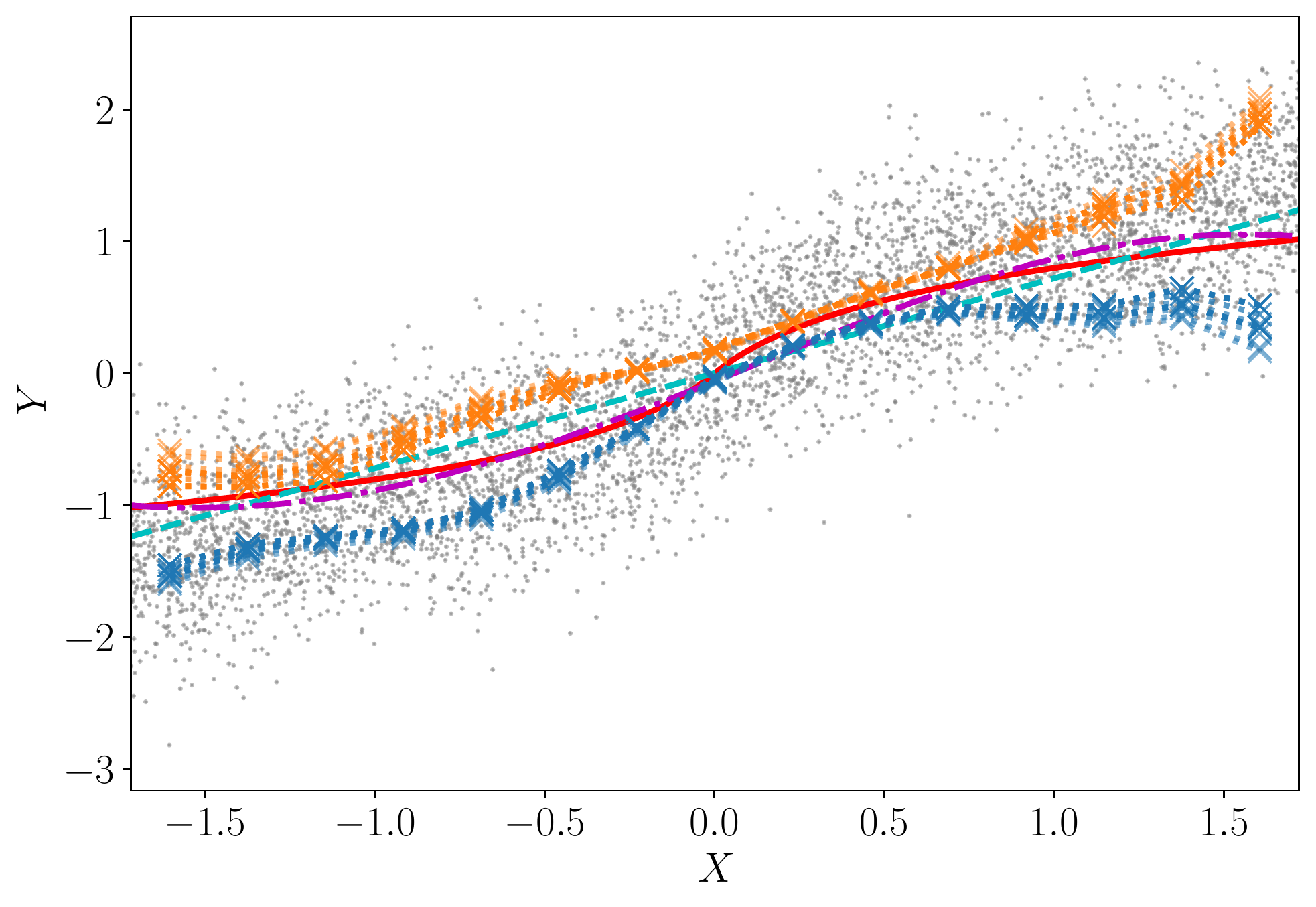}%
  \hfill
  \includegraphics[width=\fracwidth\textwidth]{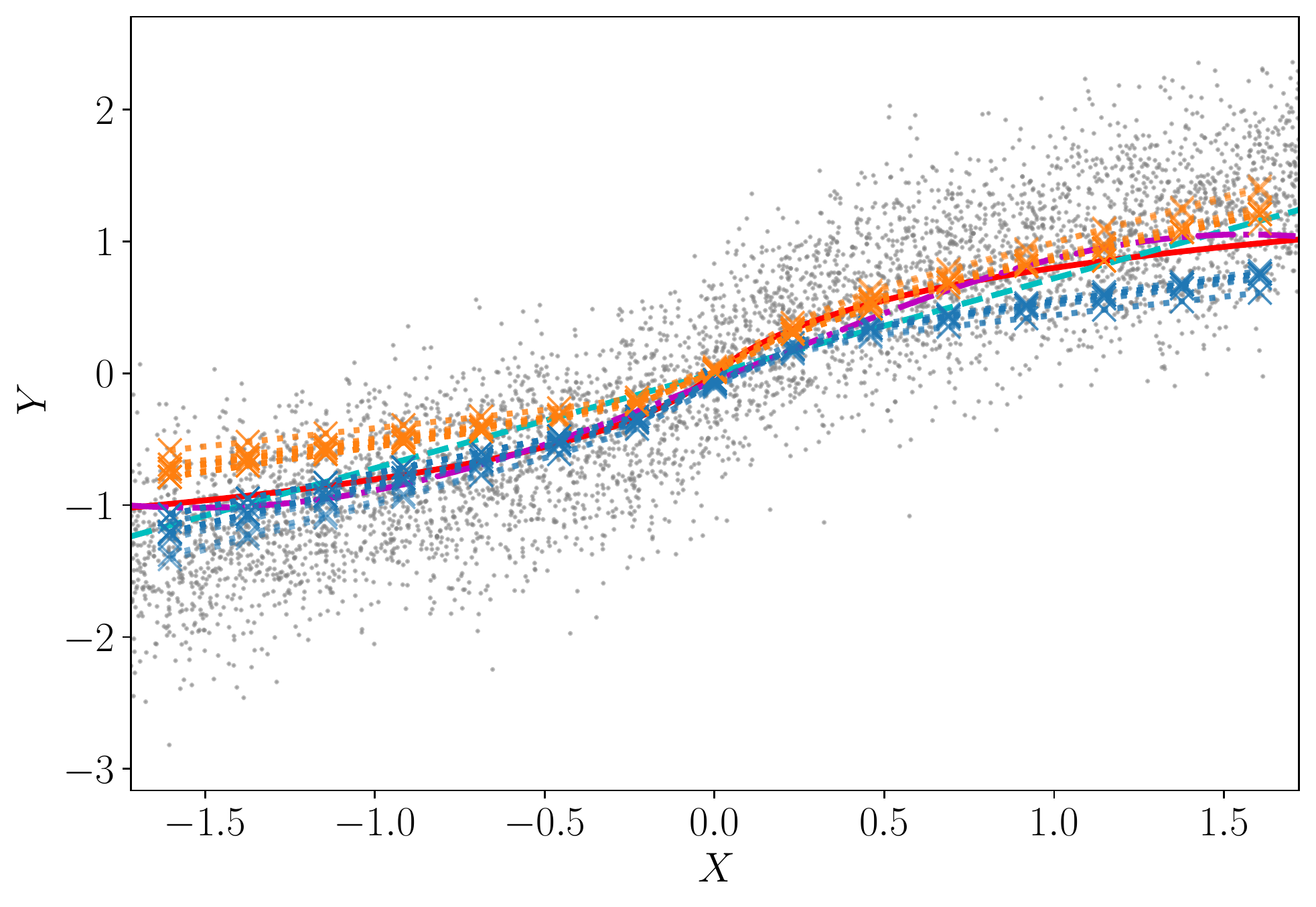}%
  \caption{Bounds for the simulated sigmoidal design.
  The true causal effect is given by a logistic function, which is well recovered by our method for different response function families (cubic polynomials, GP basis functions, and MLP basis functions).}
  \label{fig:bounds_np}
\end{figure}

\subsection{Hyperparameter Settings}
\label{app:hyperparams}

In all experiments, we fix hyperparameters $M = 20$, $L = 2$, $B = 1024$ and run SGD with momentum $0.9$ and learning rate $0.001$ for 150 rounds of the augmented Lagrangian with 30 gradient updates for each subproblem optimization.
We start with a temperature parameter $\tau=0.1$ and multiply it by $\alpha = 1.08$ in each round, capped at $\tau_{\max}=10$.
We use 7 neurons in the last hidden layer of the feed-forward neural net for MLP response functions in our synthetic setting and 9 for the expenditure data.
For GP basis functions (see Appendix~\ref{app:response}), we sample 7 basis functions for the sigmoidal design dataset (see Appendix~\ref{sec:sigmoidal}).
This set of hyperparameters did not require much manual tuning and worked for all datasets and response function families, i.e., also different dimensionality of $\theta$.
For the synthetic settings, we sample 5000 observations each.
We use 3 as the latent dimension when fitting our latent variable models.
For the tolerances, we use $\abstol=0.2$ for the synthetic settings, $\abstol = 0.1$ for the sigmoidal design (see Section~\ref{sec:sigmoidal}), $\abstol = 0.3$ for the expenditure dataset and gradually tighten $\reltol$ from $0.3$ to $0.05$ in all settings (which corresponds to the increasingly opaque lines).

\subsection{Sigmoidal Design}
\label{sec:sigmoidal}

We also evaluate our method on simulated data from a sigmoidal design introduced by \citet{chen2018optimal}, adopted by \citet{newey2003instrumental} and used in previous work on continuous instrumental variable approaches under the additive assumption as a common test case \citep{hartford2017deep,singh2019kernel, muandet2019dual}.
We show the results from KIV and our bounds for response function families consisting of cubic polynomials and neural net basis functions in Figure~\ref{fig:bounds_np}.
The observed data distribution $\hat{p}(y \given x)$ follows the true causal effect rather closely and the instrument is relatively strong in this setting, see~\citet{singh2019kernel} for details.
Therefore, the gap between our bounds is relatively narrow for a broad set of different basis functions as long as they are flexible enough to capture a sigmoidal shape.

\subsection{Expenditure Data}
\label{app:data}

We prepare the data from \citet{expenditure2000} using the same steps as \citet{gunsilius2019bounds} closely following \citet{newey2003instrumental,blundell2007semi}.
This is, we restrict the sample to households with married couples who live together and in which the head of the household is between 20 and 55 years old.
We further exclude couples with more than 2 children.
Finally, we also require the head of the household not to be unemployed.
Otherwise, the instrument, gross earnings, would not be available.
After these restrictions, we end up with 1650 observations in our dataset.
The dataset can be downloaded for free for academic purposes after creating an account.

\subsection{Small Data Regime}
\label{app:small_data}

\begin{figure}
  \centering
  \includegraphics[width=\textwidth]{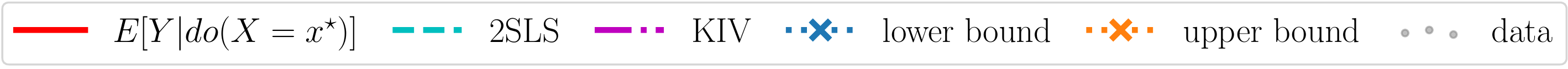}\\
  linear Gaussian settings using linear response functions\\
  \includegraphics[width=0.4\textwidth]{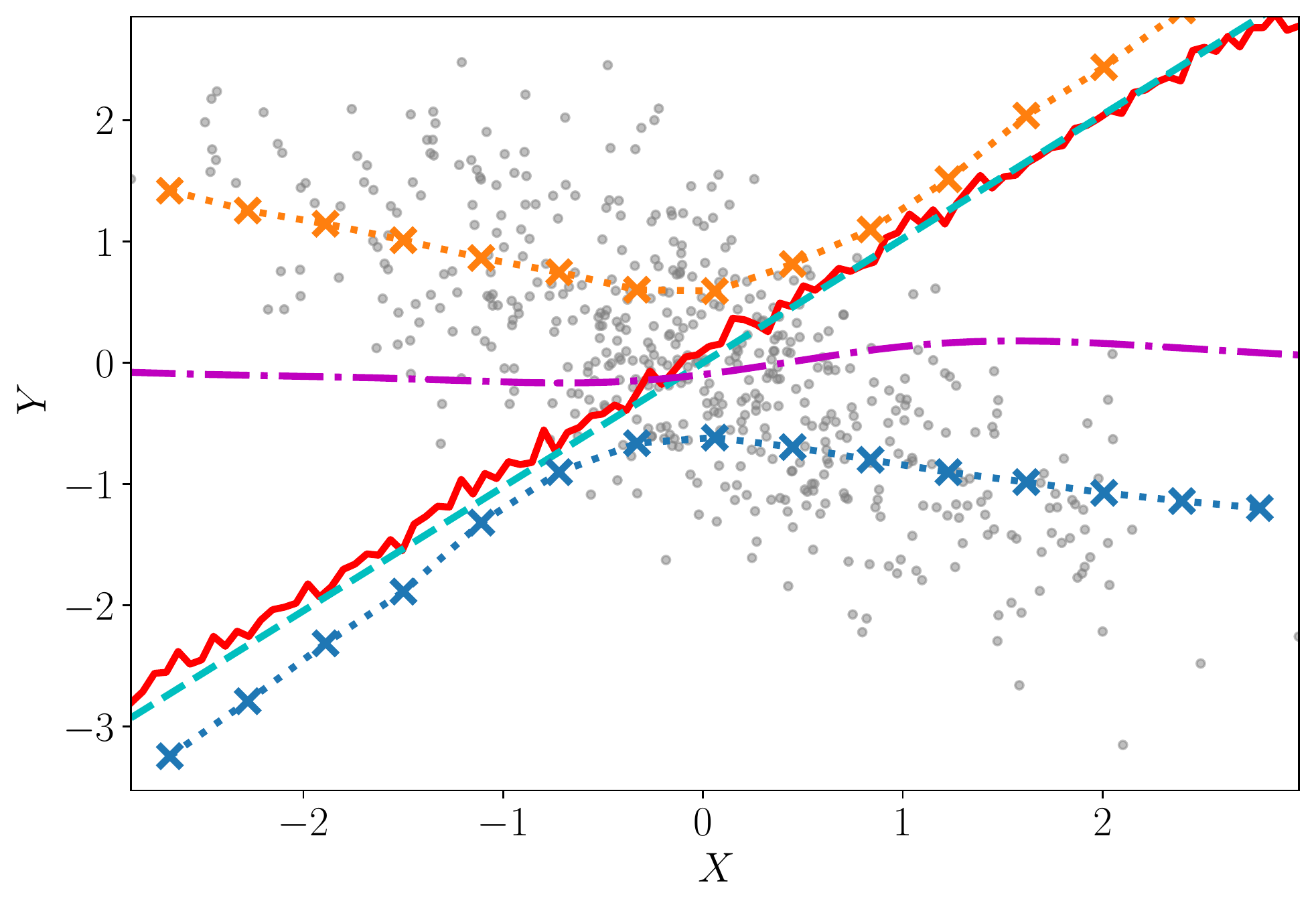}%
  \hspace{1cm}
  \includegraphics[width=0.4\textwidth]{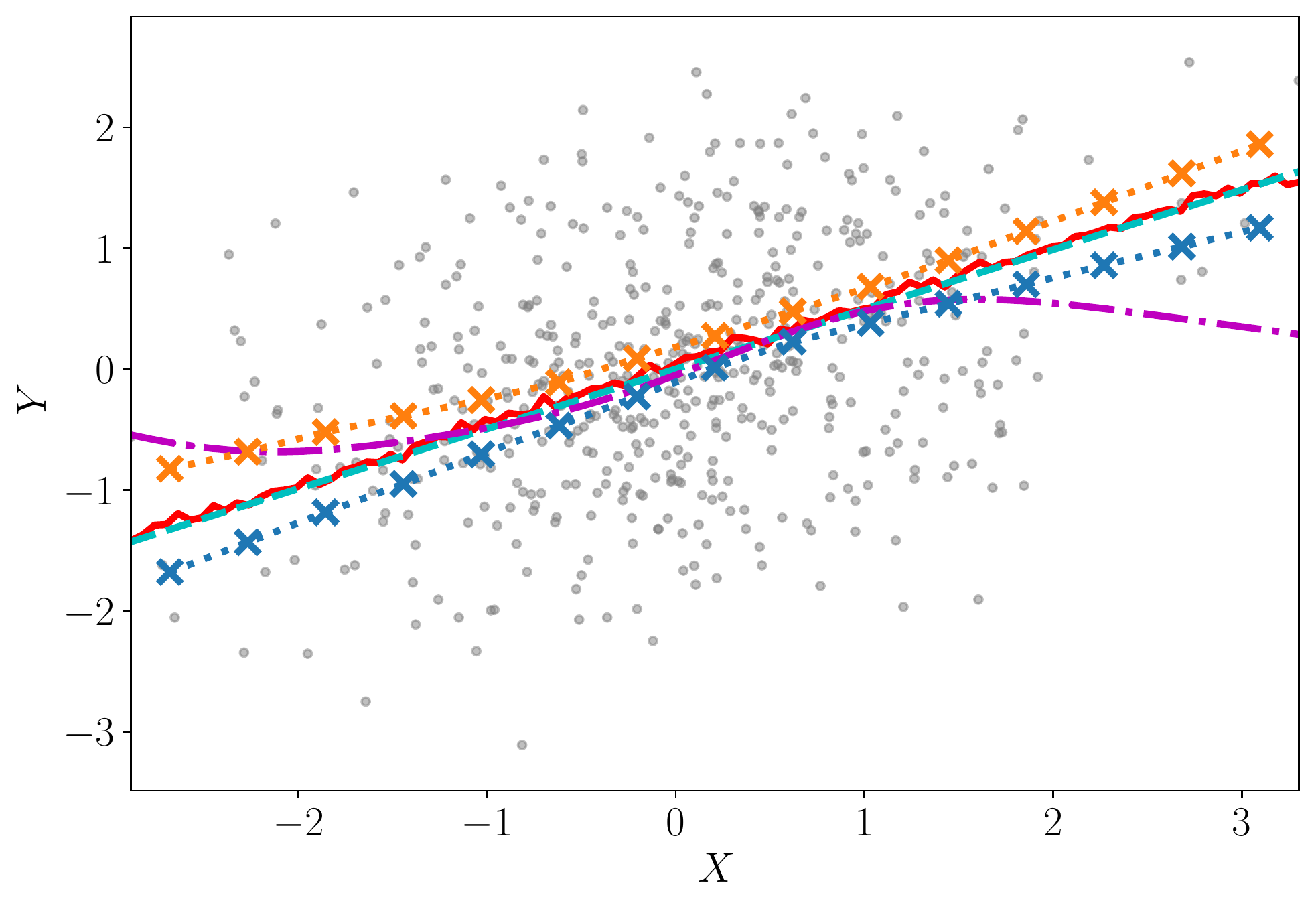}\\
  non-linear, non-additive settings using quadratic response functions\\
  \includegraphics[width=0.4\textwidth]{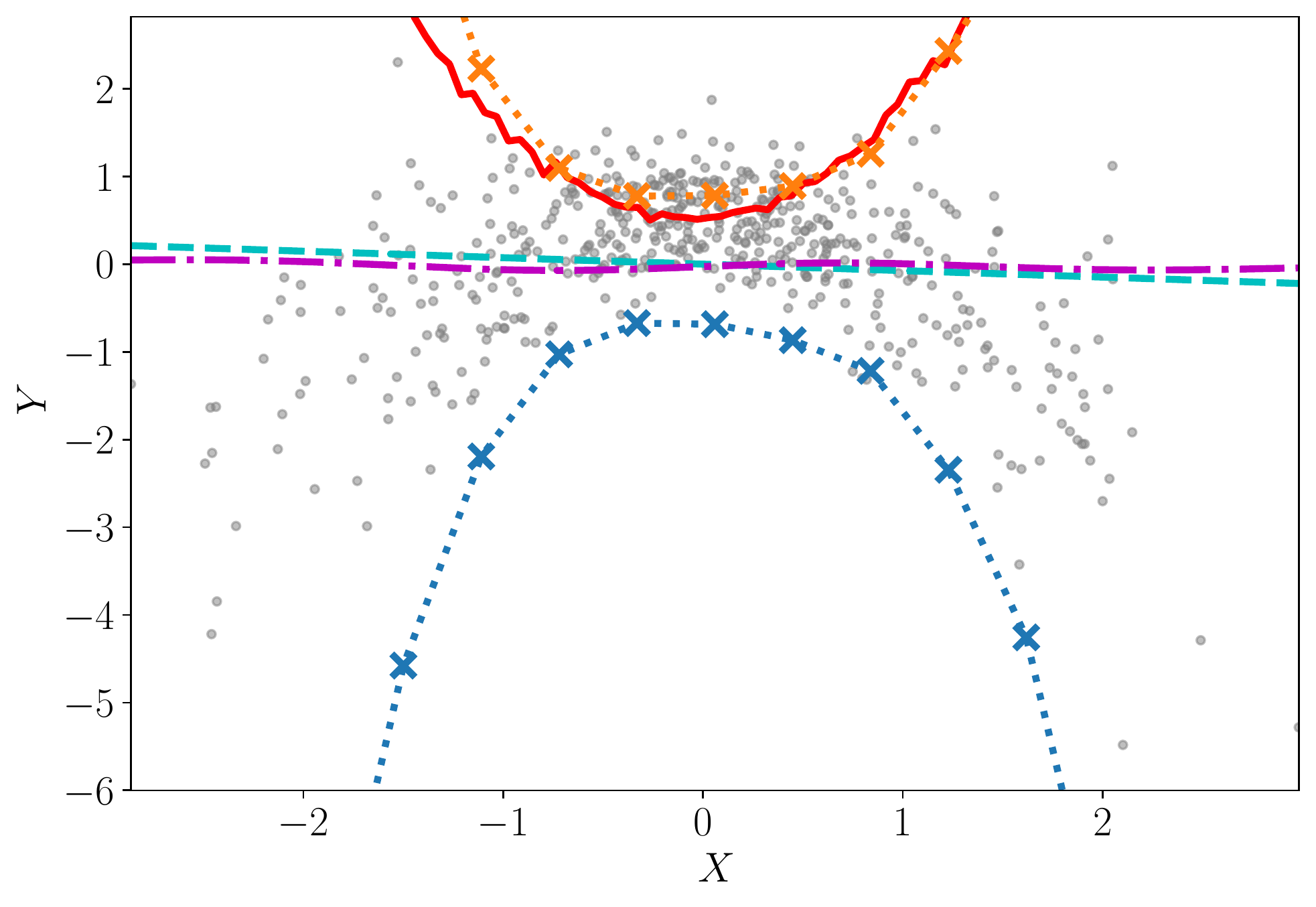}%
  \hspace{1cm}
  \includegraphics[width=0.4\textwidth]{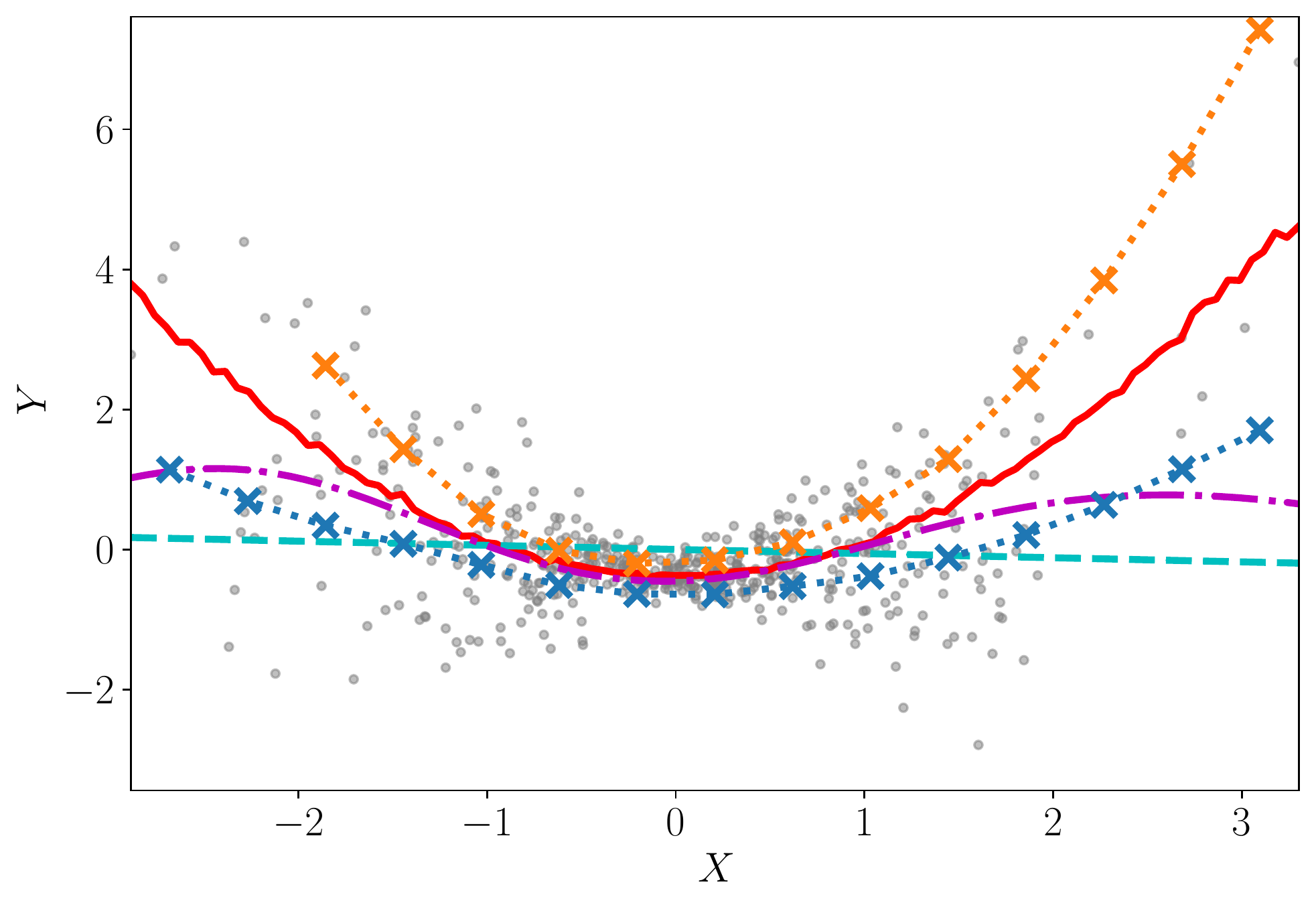}%
  \caption{Performance of our method on smaller datasets with only 500 observations.
  The left column is the strong confounding weak instrument case ($\alpha = 0.5, \beta = 3$) and the right column is the weak confounding strong instrument case ($\alpha = 3, \beta = 0.5$).}
  \label{fig:synth_small_data}
\end{figure}

\begin{figure}
  \centering
  \includegraphics[width=0.4\textwidth]{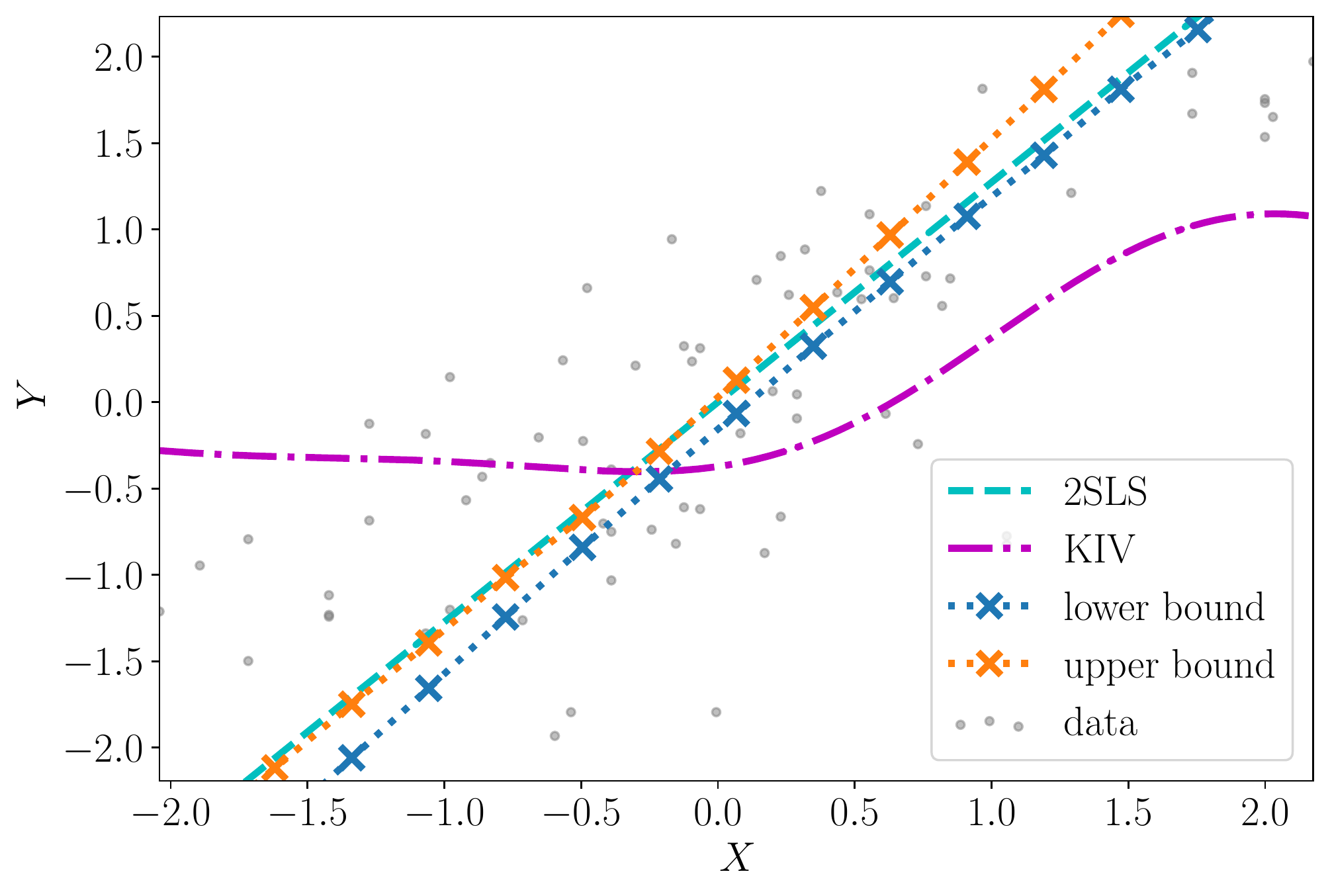}%
  \caption{Results for the small dataset from \citet{acemoglu2001colonial} with linear response functions and $M=5$ $z$-bins.}
  \label{fig:colonial}
\end{figure}

Having tested our method on datasets of size 5000 (synthetic) and 1650 (expenditure data, see Appendix~\ref{app:data}), we now evaluate how our method performs on even smaller datasets.
To this end, we first look at our synthetic settings using only 500 datapoints and correspondingly reducing the number of $z$-bins to $M=6$ in Figure~\ref{fig:synth_small_data}.
While the bounds are looser, our method can still provide useful information with relatively little data.

In addition, we ran our methods on a classic instrumental variable setting from economics, namely the dataset used by \citet{acemoglu2001colonial} on using settler mortality as an instrument to estimate the causal effect of the health of institutions on economic performance.\footnote{The dataset is freely available at \url{https://economics.mit.edu/faculty/acemoglu/data/ajr2001}.}
This dataset consists of only 70 datapoints.
Therefore, we set the number of $z$-bins to $M=5$ for this dataset.
Restricting ourselves to linear response functions, our method still gives informative bounds, which include the effect estimated by 2SLS, but does not fully include the KIV results, see Figure~\ref{fig:colonial}.

\section{KIV Heuristic for Tuning Hyperparameters}
\label{app:optimistickiv}

\begin{figure}
  \centering
  linear Gaussian setting with strong confounding and weak instrument ($\alpha=0.5$, $\beta=3$)\\
  \includegraphics[width=0.4\textwidth]{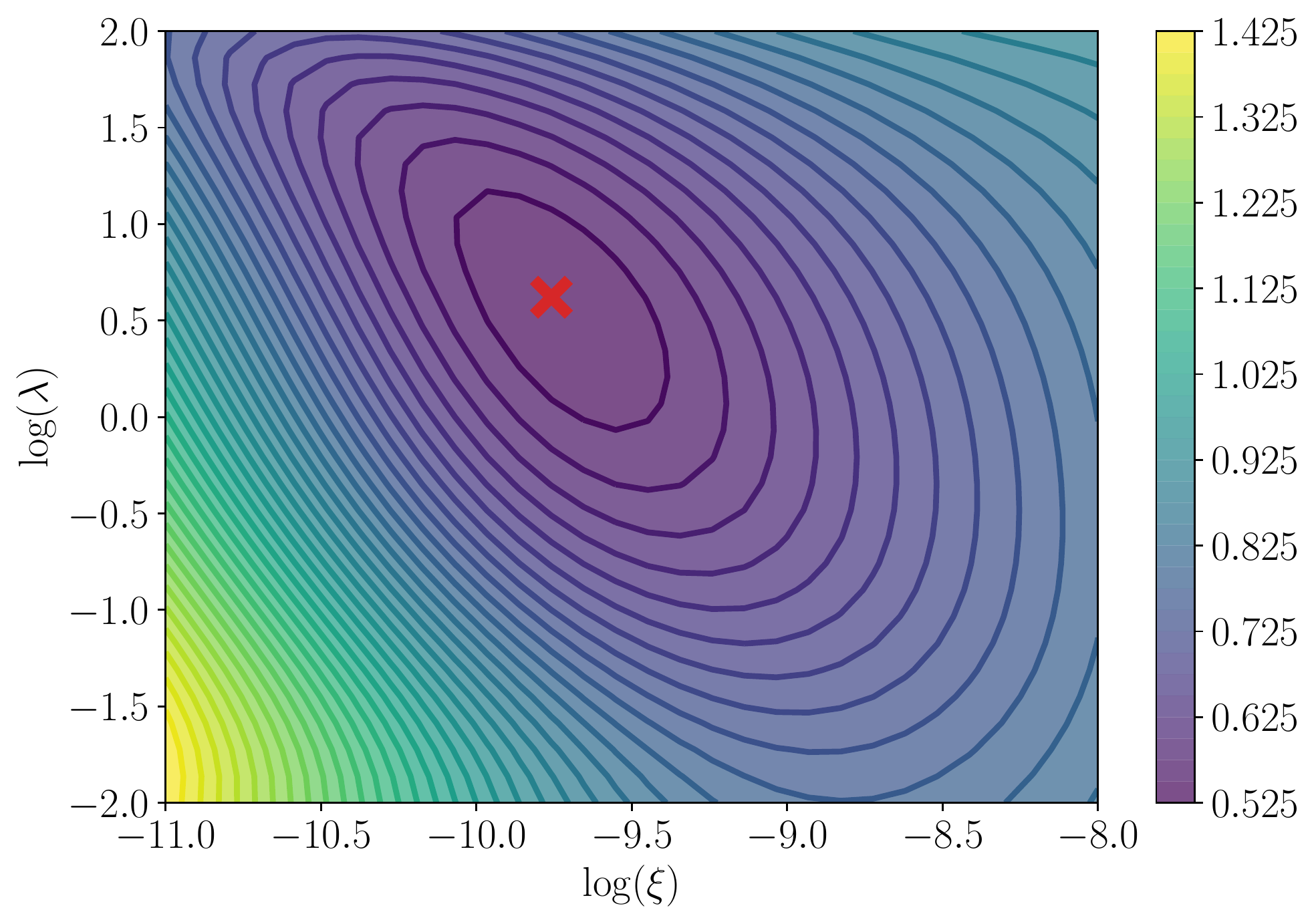}%
  \hspace{1cm}
  \includegraphics[width=0.4\textwidth]{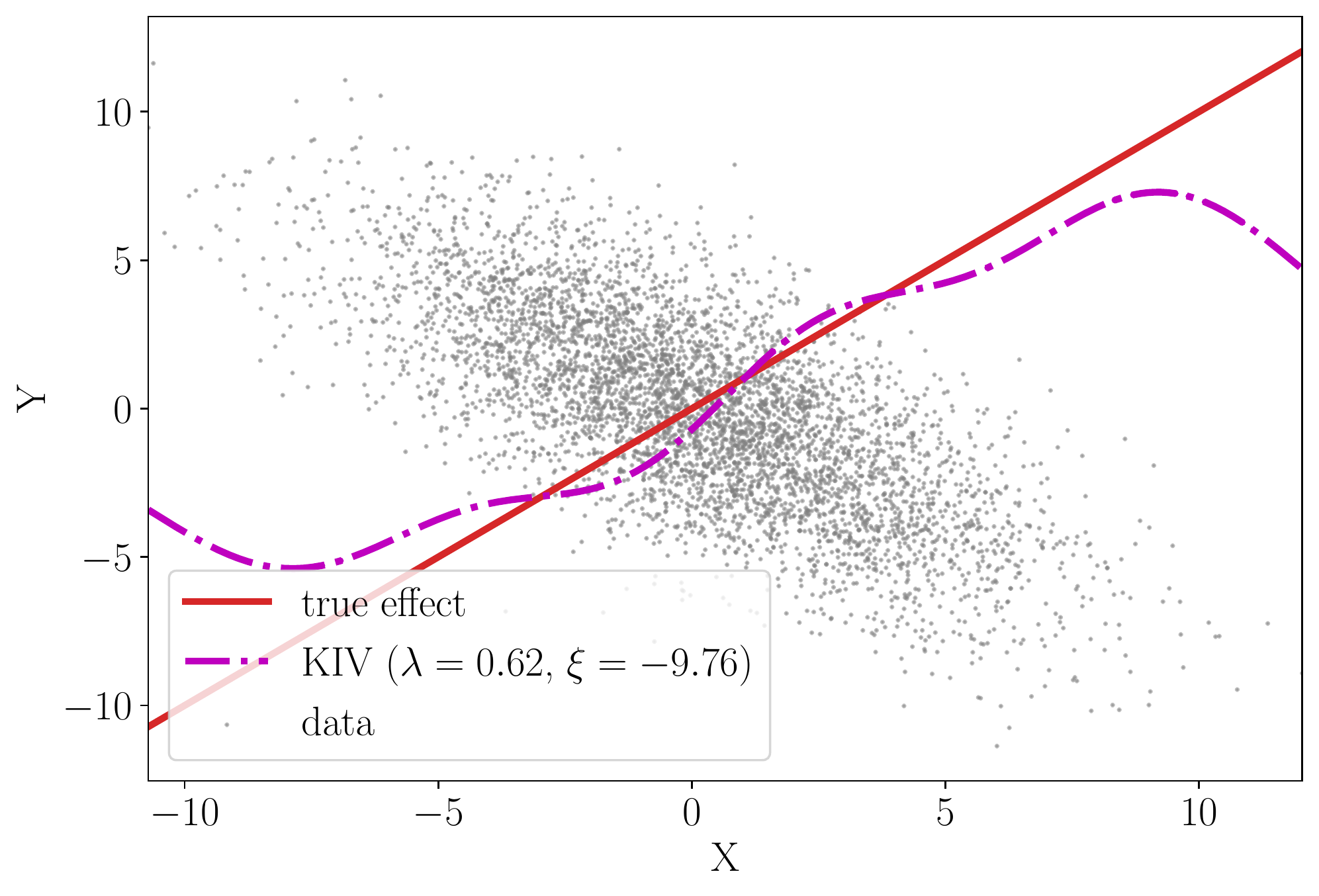}\\
  non-linear, non-additive setting with strong confounding and weak instrument ($\alpha=0.5$, $\beta=3$)\\
  \includegraphics[width=0.4\textwidth]{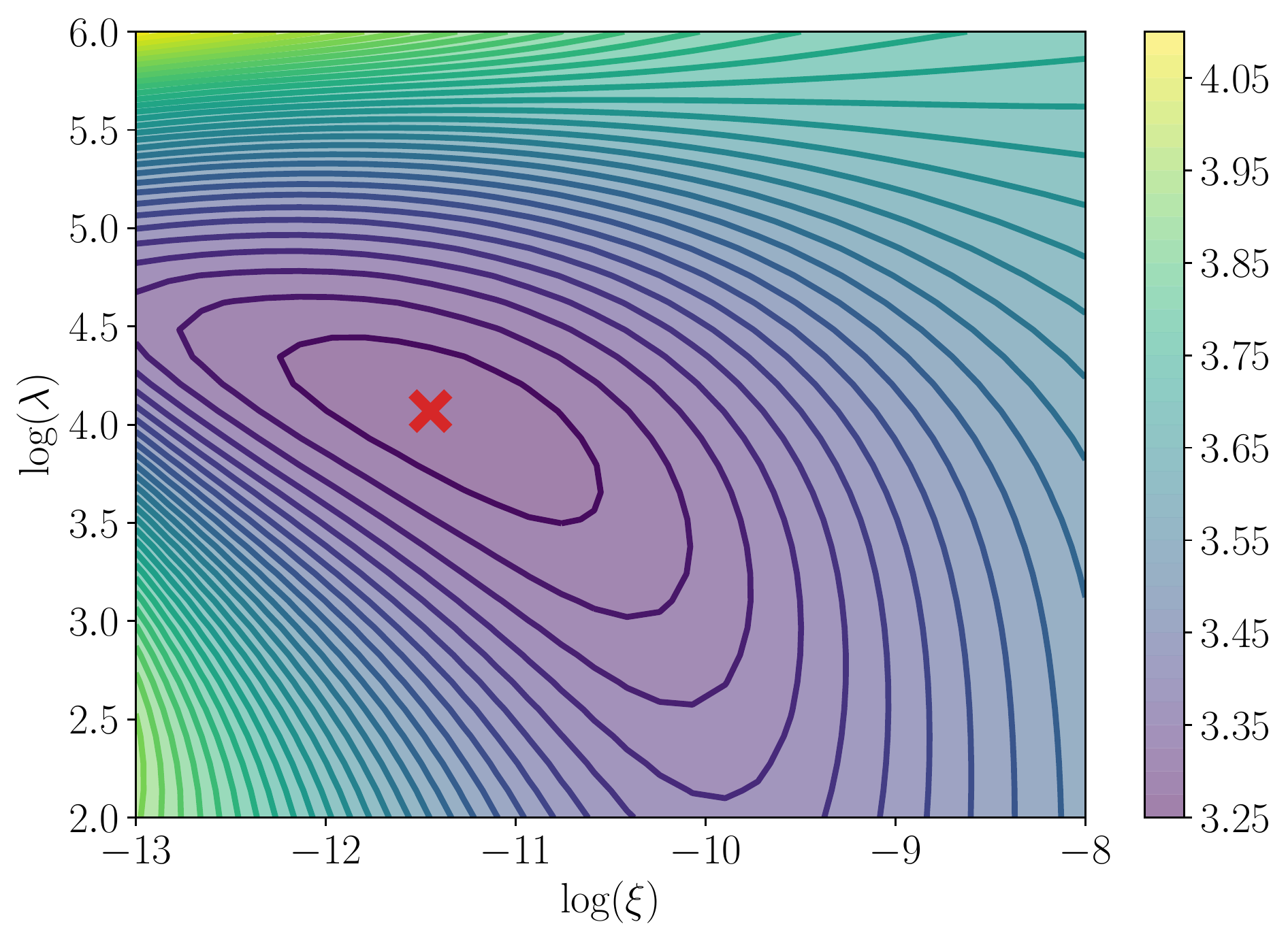}%
  \hspace{1cm}
  \includegraphics[width=0.4\textwidth]{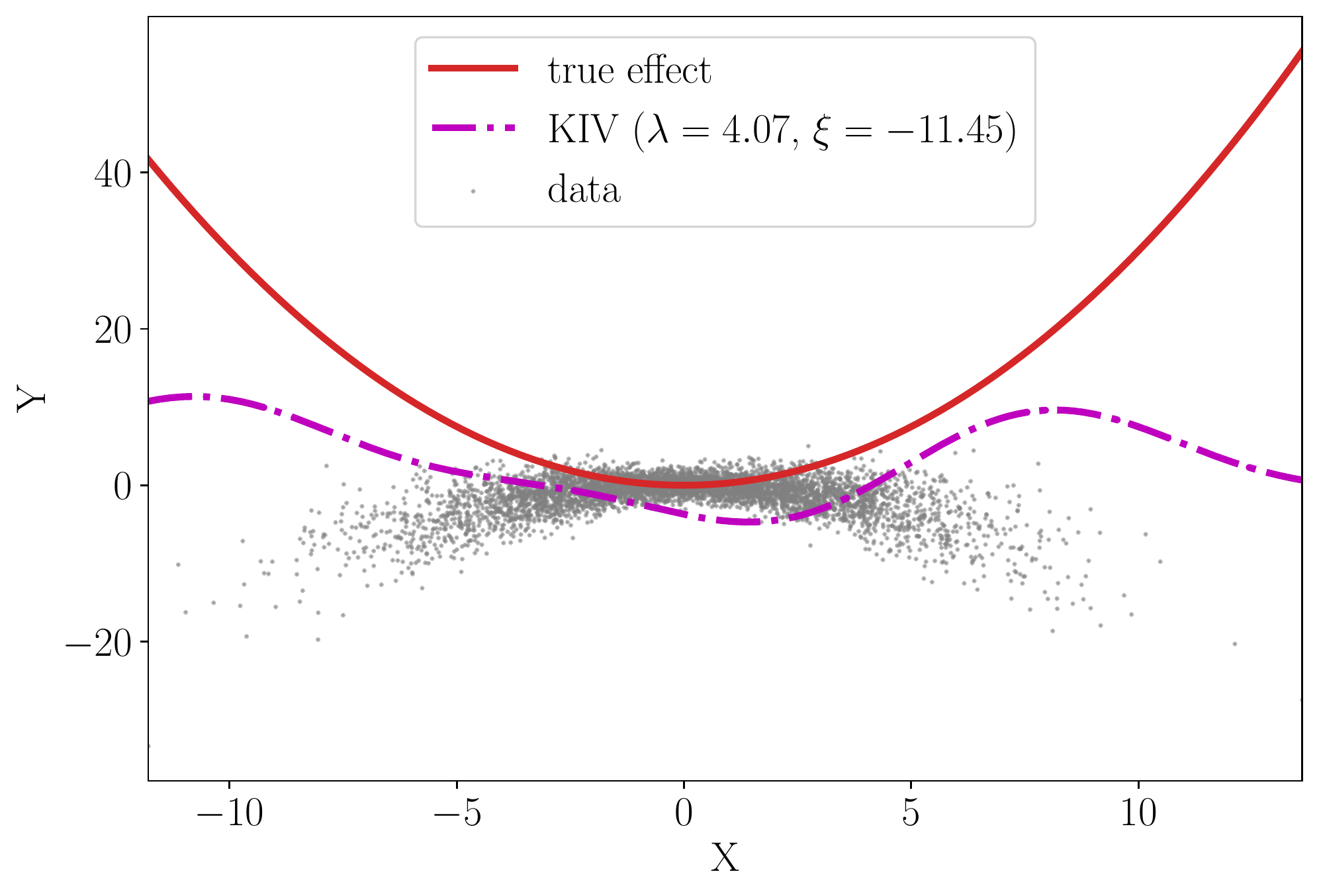}%
  \caption{We show the results of a manual hyperparameter search for KIV in the left column, where we score different settings in the two-dimensional hyperparameter space by the log of the out-of-sample mean squared error, which requires knowledge of the true causal effect.
  The red cross denotes the setting with the smallest out-of-sample mean squared error.
  In the right column, we show the KIV regression lines using the hyperparameters found in the manual search.
  The first row corresponds to the linear Gaussian setting and the second row to the non-linear, non-additive synthetic setting.}
  \label{fig:optimistickiv}
\end{figure}

We have found KIV to fail in the strongly confounded linear Gaussian setting, even though all the assumptions are satisfied, see Figure~\ref{fig:bounds_synth} (row 1).
Closer analysis of these cases showed that the heuristic that determines the hyperparameters does not return useful values in this setting.
Instead, we performed a grid search over the main hyperparameters $\lambda$ and $\xi$ \citep[see][for details]{singh2019kernel} and scored them by the out-of-sample mean-squared-error for the true causal effect (which is known in our synthetic setting).
After manual exploration of the parameter space, we found a good setting marked by the red cross in the first row on the left of Figure~\ref{fig:optimistickiv}.
Using these fixed hyperparameters for KIV instead of the internal tuning stage, we get a much better approximation of the true causal effect shown in the first row on the right of Figure~\ref{fig:optimistickiv}.
Towards the data starved regions at large and small $x$-values, KIV again reverts back towards the prior mean of zero as expected.
It is unclear at the moment, however, how to set such hyperparameter values without access to the true causal effect.
Our point here is that in principle there is a setting with acceptable results, although even then it is not clear how much of it is a coincidence based on looking at many possible configurations.

We performed a similar manual analysis for the non-linear, non-additive synthetic setting with strong confounding, in which off-the-shelf KIV fails as well, see Figure~\ref{fig:bounds_synth} (row 3).
Note that this setting does not satisfy the assumptions of KIV, because of the non-additive confounding.
Again, we do manage to find hyperparameters that locally minimize the out-of-sample mean-squared-error shown in the second row on the left of Figure~\ref{fig:optimistickiv}.
However, the resulting regression of the causal effect does not properly capture the true effect as shown in the second row on the right of Figure~\ref{fig:optimistickiv}.

\end{document}